\pdfoutput=1
\documentclass[10pt,twocolumn,letterpaper]{article}

\usepackage[pagenumbers]{cvpr} 

\usepackage[dvipsnames]{xcolor}

\definecolor{cvprblue}{rgb}{0.21,0.49,0.74}
\usepackage[pagebackref,breaklinks,colorlinks,citecolor=cvprblue]{hyperref}

\def\paperID{*****} 
\def\confName{CVPR}
\def\confYear{2024}

\title{Stable Diffusion Reference Only: Image Prompt and Blueprint Jointly Guided Multi-Condition Diffusion Model for Secondary Painting}

\author{Hao Ai \\
Beihang University\\
{\tt\small aihao@buaa.edu.cn}
\and 
Lu Sheng*\\
Beihang University\\
{\tt\small lsheng@buaa.edu.cn}
}

\begin{document}
\maketitle
\begin{abstract}
    Stable Diffusion and ControlNet have achieved excellent results in the field of image generation and synthesis. However, due to the granularity and method of its control, the efficiency improvement is limited for professional artistic creations such as comics and animation production whose main work is secondary painting.
    In the current workflow, fixing characters and image styles often need
    lengthy text prompts, and even requires further training through
    TextualInversion, DreamBooth or other methods, which is very complicated and
    expensive for painters.
    Therefore, we present a new method in this paper, Stable Diffusion Reference Only, a images-to-image self-supervised model that uses only two types of conditional images for precise control generation to accelerate secondary painting.
    The first type of conditional image serves as an image prompt, supplying the necessary conceptual and color information for generation.
    The second type is blueprint image, which controls the visual structure of the generated image. It is natively embedded into the original UNet, eliminating the need for ControlNet.
    We released all the code for the module and pipeline, and trained a controllable character line art coloring model at \href{https://github.com/aihao2000/stable-diffusion-reference-only}{https://github.com/aihao2000/stable-diffusion-reference-only}, that achieved state-of-the-art results in this field.
    This verifies the effectiveness of the structure and greatly improves the production efficiency of animations, comics, and fanworks.
\end{abstract}
\section{Introduction}
\label{sec:intro}

Secondary Painting refers to works that primarily involve modification,
association, and derivation based on a given concept. It is the largest work in
comics, animations, fanworks. With the popularity of comics and animations,
Secondary Painting tasks are also increasing. A common task is to draw new
poses for specific characters. Especially for animations with high frame rates,
this is a very repetitive and workload-heavy task.

However, this work is not currently being targeted as a major exploration. In
recent years, breakthroughs in image generation models have been mainly driven
by text-guided diffusion models. Stable Diffusion\cite{Rombach_2022_CVPR}, the
latest text-to-image model, can efficiently understand and extract text
semantics to generate images of related concepts.
ControlNet\cite{zhang2023adding}, in text-to-image generation, controls the
visual structure of the generated image by copying the
UNet\cite{ronneberger2015unet} model and adding conditional images. However,
all these methods are text-driven and have a limitation in being able to apply
the concepts extracted directly from new images to online generation. It is
difficult to describe a specific character or image style in words, and Stable
Diffusion based on CLIP's\cite{radford2021learning} text transformer model has
a maximum word limit of 77. Professional training and hyperparameter tuning is
a high-cost thing for painters.

Recently, there has been some additional work on the text-to-image pipeline,
such as ControlNet Reference Only or IP-Adapter\cite{ye2023ipadapter}, which
have the ability to implement secondary painting with another ControlNet. But
in fact, its generalization to new characters is not good, and it is not
qualified for the task of coloring manga characters.

Existing methods rely on Stable Diffusion after text-to-image training. We
speculate that existing methods perform poorly on untrained characters because
they do not change the probability distribution of generated images originally
trained with text-to-image. Text-based control is less precise compared to
image-based control. On this basis, the text-aligned features obtained by
training only the image encoder are too rough to be general enough. Therefore,
to achieve a comprehensive images-to-image model for secondary painting, it is
essential to modify the initial training method. The related modules should be
trained together in order to obtain the most common image features. This will
enable the learning of a formula for extracting color and shape information
from the reference image, which can then be used to populate the blueprint.

Threfore, in this paper, we focus on the images-to-image pipeline and present a
new model called Stable Diffusion Reference Only. This is a self-supervised
model supports multi-condition control. For painters, they can get
plug-and-play secondary painting products so that generating images of a
specific style and character will no longer require lengthy text prompts or
professional fine-tuning. Stable Diffusion Reference Only only needs two image
conditions. The first is image prompt, which provides the conceptual and color
information needed to generate the image. For example, it could be an
illustration of a character design. The second one is blueprint image, which
controls the visual structure of the generated image. It has similar
functionality to the condition image of ControlNet but does not require the
same level of resource cost and extra training. It is introduced in unified
training to reduce training difficulty and improve the quality of results.

We follow the basic structure of Stable Diffusion and reuse its
VAE\cite{kingma2022autoencoding} to perform multi-condition guided diffusion in
the latent space. The image prompt is encoded using the vision
transformer(ViT)\cite{dosovitskiy2021image} and applied to the UNet backbone
via the cross-attention\cite{vaswani2023attention} mechanism. ControlNet's
ablation study mentioned that when prompts are sufficient, the lighter
ControlNet can achieve the same effect. Therefore, we feel free to add the
blueprint input to the original UNet, and add a simple convolutional network to
it to affect the Q matrix of the attention mechanism. This is very intuitive.
Each patch of the image looks at the image prompt to ask how it should be
colored.

Based on the above structure, a multi-condition guided diffusion model can be
constructed, which takes the image prompt and blueprint image as inputs to
generate the target image. This model functions as a native image-to-image
model, allowing for self-supervised training. To train the model, we utilize
CLIP to identify similar image pairs. In this approach, one image serves as the
image prompt, another image represents the desired result image, and an
adaptive threshold and color inversion are applied to create the blueprint
image.

A notable advantage of this approach is that we train under a unified loss
function to obtain a model with stronger generalization and greatly reduce the
cost of training and inference. We verified the feasibility of this structure
in the most critical task of character line art coloring. Painters only need to
provide any character design to color the new line art. This greatly
facilitates secondary painting and speeds up the production process of comics,
animations and fanwork.

In sum, out work makes the following contributions:

(i) We present a new UNet structure, which perfectly corresponds to the requirements of multi-condition diffusion and secondary painting.

(ii) Based on the new UNet, we designed an native images-to-image model and a corresponding self-supervised learning method. All modules are trained uniformly to obtain a model with stronger generalization.

(iii) Painters can obtain an plug-and-play secondary painting product that utilizes two conditional images to accurately control the generated image for the line art coloring task, eliminating  the need for professional model fine-tuning.

(iv) Finally, we released the code of all modules, pipelines, and an automatic coloring pre-trained model.
\section{Related Work}
\label{sec:formatting}

Stable Diffusion Model is a text-controlled image generation model that builds
upon the efficiency of the diffusion structure, as verified by DDPM (Denoising
Diffusion Probabilistic Model) \cite{radford2021learning}, in the field of
image generation. The Stable Diffusion Model leverages the latent space to
significantly reduce the computational resources required for training and
inference at high resolutions. Additionally, it incorporates the text encoder
from CLIP (Contrastive Language-Image Pretraining) to enable text-to-image
generation. However, when it comes to generating new image concepts, such as
specific characters, the existing text-to-image workflows are not sufficient.
The model lacks the native ability to directly exploit these new image
concepts. Describing images using general words is often challenging, leading
painters to rely on techniques like DreamBooth \cite{ruiz2023dreambooth} and
Textual Inversion \cite{gal2022image} for fine-tuning their work.
Unfortunately, these techniques come with high learning costs and computing
power requirements for creators. The principle of original image-to-image
pipeline is to use a hint image to replace the initial noise image, which can
simply fix the visual structure of the image, but it cannot achieve the effect
of providing new concepts.

ControlNet. To be precise, what is currently popular is ControlNet-Self, a copy
diffusion UNet that uses convolutional layers to connect new conditional inputs
and the output of each layer as an encoder to control the generated image
structure in the text-to-image workflow. The author discussed its differences
with other versions: ControlNet-Lite, ControlNet-MLP in the Ablation Study.
ControlNet-Self is equivalent when there is a sufficient text prompt, but works
better when there is not. However, limited by the workflow of text-to-image, if
painters want to fix the picture structure and control the specific content,
they still have to rely on text. For example, in line art coloring tasks, the
rich colors are difficult to describe using words. Then the creators have to go
back to expensive training and hyperparameter tuning. ControlNet recently
provided a similar work - ControlNet Reference Only, which can provide a simple
object image and use Multi-ControlNet to achieve the generation of specific
objects with other ControlNet, hoping to solve the limitations of text-to-image
in the field of secondary painting. However, multiple ControlNets have not been
jointly trained, but simply add the injected features, which has great
uncertainty. And based on the structure of ControlNet-Self, it relies very much
on the UNet of the base model. It is similar to a classifier based on UNet
features of the base model. For new features such as new characters, the
performance will be very poor.

IP-Apater is the latest image prompt work recently. It adds an image
conditional embedding mechanism very similar to ours on the text-to-image
model. And compatible with ControlNet. Therefore, the line art coloring can be
implemented by combining ControlNet and a image encoder. However, its
generalization to unknown data is not good enough, especially for the coloring
of new manga characters. We speculate that the reason is that the image encoder
trained on the basis of text-to-image obtains text-aligned image features that
are too rough to achieve universal image feature representation. Therefore, its
painting style is does not match and the color is not accurate. This solution
cannot learn a general formula to use the elements in the image prompt to fill
in the corresponding positions of the blueprint.

\section{Method}

We follow StableDiffusion to perform conditional guided diffusion in latent
space. In order to achieve multiple conditions guidance and unified training
goals, we modified the structure of origin UNet. During training, we use
pre-trained models as much as possible to reduce training difficulty, and use
CLIP and ISNet\cite{qin2022highly} to obtain high-quality training data to
train a model with higher generalization. Finally, We achieved state-of-the-art
results in the most important task in the field of comics and animation
creation: line drawing coloring.
\subsection{Conditioning Mechanisms}

Similar to Stable Diffusion and ControlNet, $p_\theta$ is recorded as the
back-diffusion posterior probability distribution of neural prediction, T is
the total number of diffusion steps, t is a certain diffusion step, $c_p$ is
image prompt, $c_b$ is blueprint image.

Stable Diffusion Reference Only is in principle capable of modeling conditional
distributions of the form $p(z|c_p,c_b)$.
\[
  p_\theta(z_{0:T}|c_b,c_p)=p_\theta(z_T)\prod_{t=1}^T p_\theta(z_{t-1}|z_t,c_p,c_b)
\]
This can be implemented with a conditional denoising autoencoder
$\epsilon_\theta(z_t,t,c_p,c_b)$.

In order to introduce blueprint images in the unified training, we add a new
conditional input to the original UNet of Stable Diffusion. Based on the
results of the ablation study of ControlNet mentioned above. ControlNet-Lite (a
lighter version of ControlNet) is as effective as ControlNet when conditions
are sufficient. Therefore we feel free to use lightweight embedding methods:
just use a tiny network $\mathcal{E}_\theta (\cdot)$ of four convolution layers
with $4 \times 4$ and $2 \times 2$ strides (activated by ReLU, channels are 16,
32, 64, 128, trained jointly with the full model) to encoder image-space
conditions $c_b$ into feature maps with
\[
  c_b'=\mathcal{E}_\theta(c_b)
\]
where $c_b'$ is the converted feature map. This network convert $512 \times
  512$ blueprint image conditions to $64 \times 64$ feature maps.

We enhance the underlying UNet backbone by using a cross-attention mechanism.
We use an image encoder $\tau_\theta$ to extract image concept information
$\tau_{\theta}(c_p)\in\mathbb{R}^{M\times\hat{d}_{\tau}}$, which is then mapped
to the intermediate layers of the UNet via a across-attention layer
implementing $Attention(Q,K,V)=softmax(\frac{QK^T}{\sqrt{d}})\cdot V$, with
\begin{flalign*}
   & Q=W_Q^{(i)}\cdot\varphi_i(z_t,\mathcal{E}_\theta(c_b)), \\
   & K=W_K^{(i)}\cdot\tau_\theta(c_p),                       \\
   & V=W_V^{(i)}\cdot\tau_\theta(c_p).
\end{flalign*}
Here,  $\varphi_i(z_t+\mathcal{E}_\theta(c_b))\in \mathbb{R}^{N\times
    d_\epsilon^i}$ denotes a (flattened) intermediate representation of the UNet
implementing $\epsilon_\theta$ and $W_V^{(i)} \in \mathbb{R}^{d\times
    d_\epsilon^i},W_Q^{(i)}\in\mathbb{R}^{d\times d_\tau}\ \&\
  W_K^{(i)}\in\mathbb{R}^{d\times d_\tau}$ are learnable projection matrices.

Based on the above conditional mechanism,
\[
  \mathcal{L}=\mathbb{E}_{z_0,t,c_p,c_b,\epsilon\sim\mathcal{N}(0,1)}\Big[\|\epsilon-\epsilon_\theta(z_t,t,\tau_\theta(c_p),\mathcal{E}_\theta(c_b))\|_2^2\Big]
\]
where both $\tau_\theta$, $\mathcal{E}_\theta$ and $\epsilon_\theta$ are
jointly optimized via above equation.

Finally, similar to Stable Diffusion, we also use a
VAE\cite{kingma2022autoencoding} to project the image into the latent space
under the above conditional mapping.

In summary, the entire model is shown in the figure \ref{fig:Stable Diffusion
  Reference Only}. The user uses two condition images to control the diffusion,
and one image serves as the image prompt, providing sufficient conceptual
information required to generate the picture. Another image serves as a
blueprint to control the visual structure of the generated image. In the
automatic coloring of characters in line arts, the image prompt is the
character reference image, and the blueprint is the line art that needs to be
colored. Blueprint can be easily extended to other forms such as sketches,
pose, scribble, etc.
\begin{figure}
  \centering
  \includegraphics[width=0.8\linewidth]{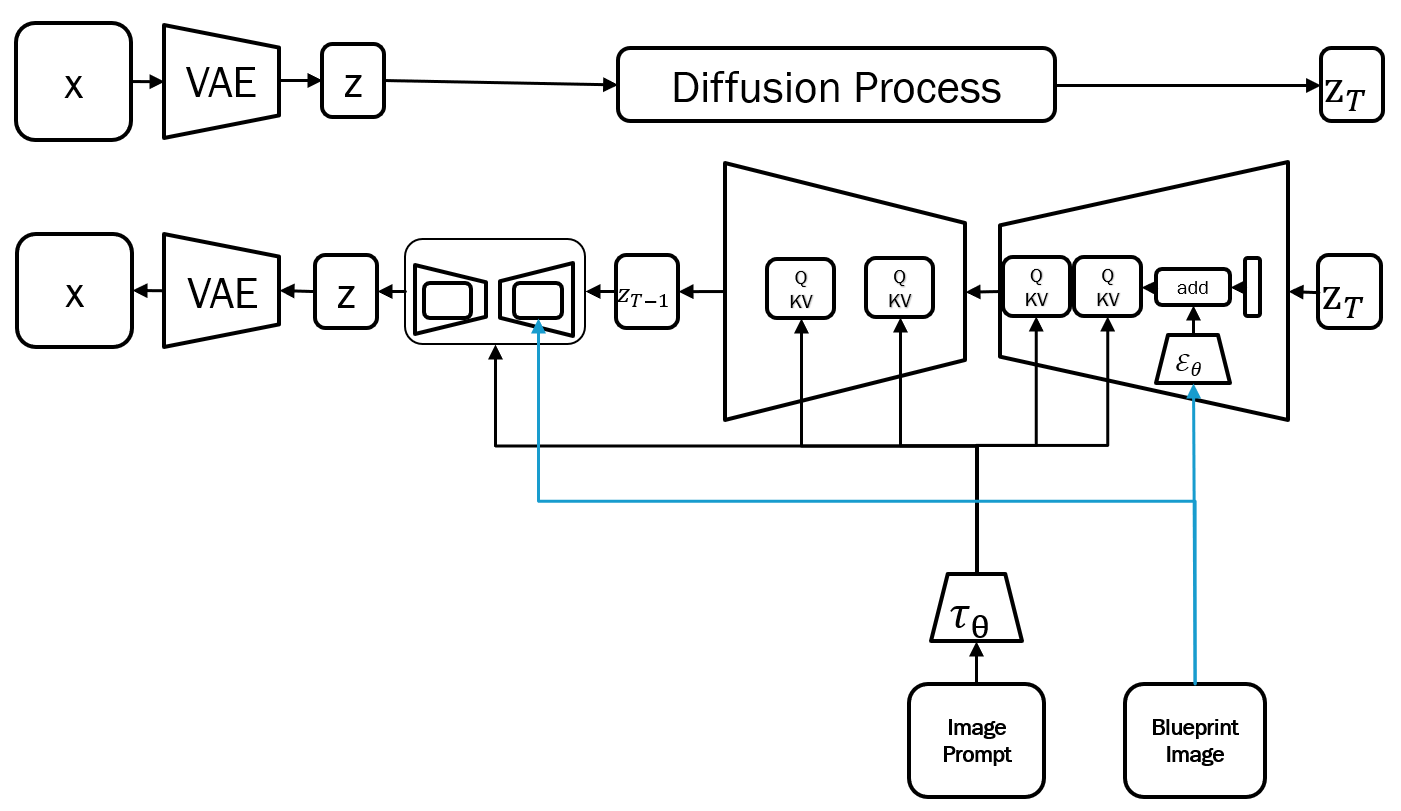}
  \caption{Stable Diffusion Reference Only}
  \label{fig:Stable Diffusion Reference Only}
\end{figure}
\subsection{Data Selection and Augmentation}

The main audience for secondary painting is in the field of animation and
comics, so our training data are all animation character pictures. We use a
private dataset of mostly AI-generated anime images. Line art coloring often
does not include background, and does not have great versatility, which will
increase the difficulty of train and bring more random effects. Therefore we
use ISNet for data augmentation to remove dirty data or background and only
retain the characters. Cleaning the data will result in a large amount of empty
entries, which are essentially completely transparent and do not contain any
characters. We use the grayscale image after the adaptive threshold method of
the OpenCV library to determine the empty images to further clear these new
dirty data. We use CLIP to find 1,000,000 pairs of images with a similarity
greater than $90\%$. Finally, use the adaptive threshold method of the OpenCV
library or the method generated by the condition of ControlNet to get the
blueprint to obtain the final training data.

\subsection{Train}

We choose UNet and VAE of Stable Diffusion 2-1 as the initial part of the
entire model. Some of the weights processed by blueprint are initialized with
0, and the rest can be simply migrated from Stable Diffusion's UNet. The Image
encoder uses the CLIP Vision Model of OpenAI clip-vit-large-patch14, which
extracts features in the 1024 dimension and perfectly matches the context
dimension of Stable Diffusion 2-1.

Training on four A100-40G. Following the training parameters of StableDiffusion
2, the learning rate uses a scheduler that wraps to the target value and then
keeps constant. Optimizer is AdamW\cite{loshchilov2019decoupled}. First, the
batch size is 100, the learning rate is 5e-6, and 100,000 steps are trained at
a resolution of $256 \times 256$. Then use batchsize as 40, learning rate as
2e-6, and train for 850,000 steps at $512\times512$. Finally, fintune is
performed on $768 \times 768$.

\section{Experiments}
\subsection{Experimental Settings}

The sampler is UniPCMultistepScheduler\cite{zhao2023unipc}. The number of
inference steps is 20.
\subsection{Qualitative Results}
We percent qualitative results in Fig \ref{tab:Stable Diffusion Reference Only:
  Automatic Coloring Test} and surprisingly find that the model can generalize to
the situation where the reference image and line art have different characters,
a feature not found in the training data. Therefore it can realize the function
of anime character style transfer.
\begin{table}
  \centering
  \begin{tabular}{@{}p{0.26\linewidth}p{0.26\linewidth}p{0.26\linewidth}@{}}
    \toprule
    Prompt                                             & Blueprint                                             & Result                                             \\
    \midrule
    \includegraphics[width=\linewidth]{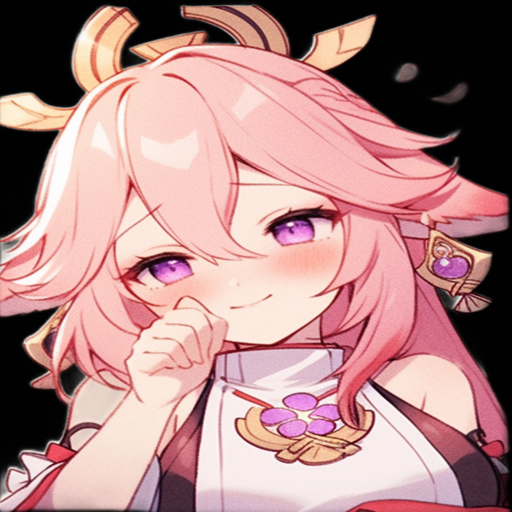}  & \includegraphics[width=\linewidth]{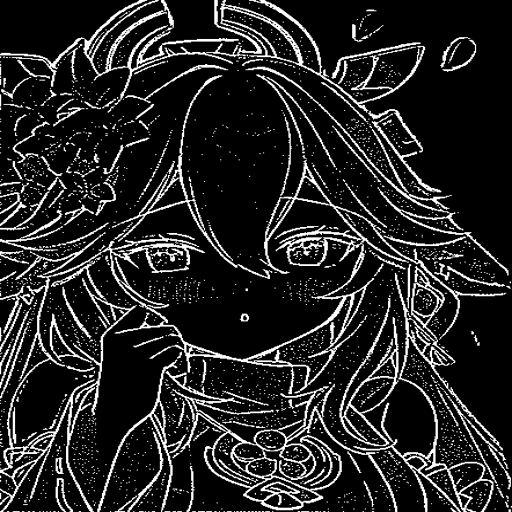}  & \includegraphics[width=\linewidth]{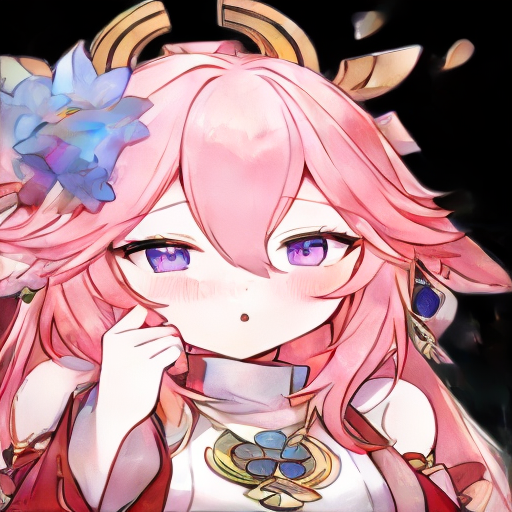}  \\
    \includegraphics[width=\linewidth]{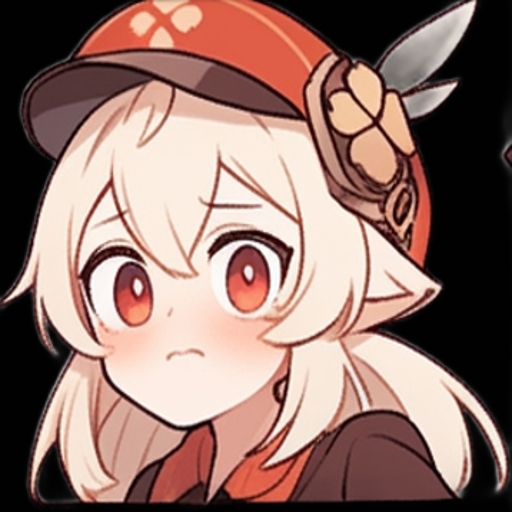}  & \includegraphics[width=\linewidth]{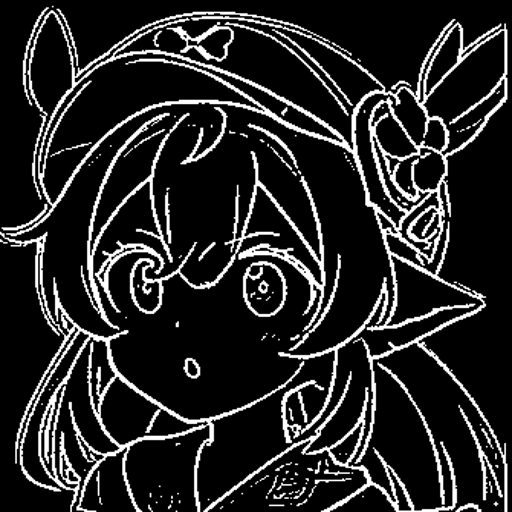}  & \includegraphics[width=\linewidth]{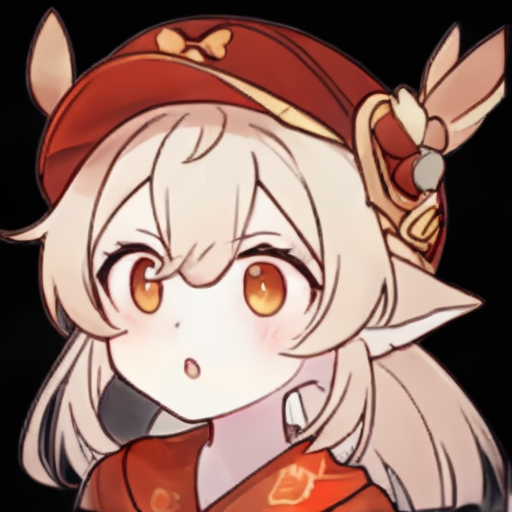}  \\
    \includegraphics[width=\linewidth]{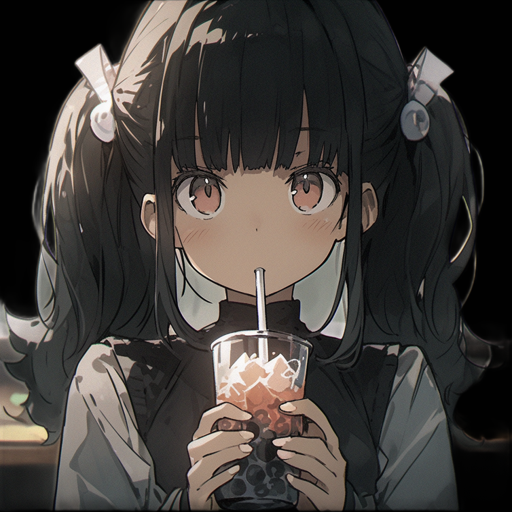}   & \includegraphics[width=\linewidth]{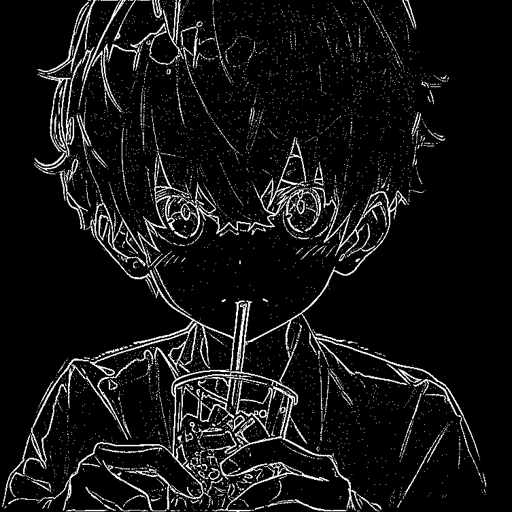}   & \includegraphics[width=\linewidth]{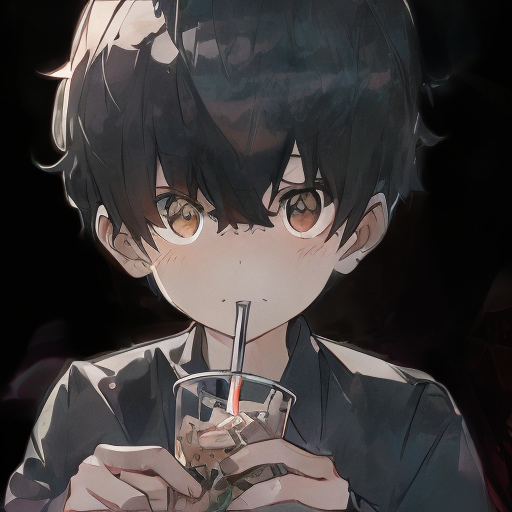}   \\
    \includegraphics[width=\linewidth]{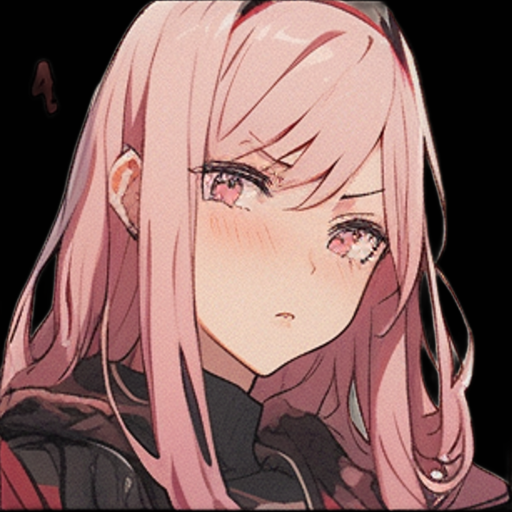}   & \includegraphics[width=\linewidth]{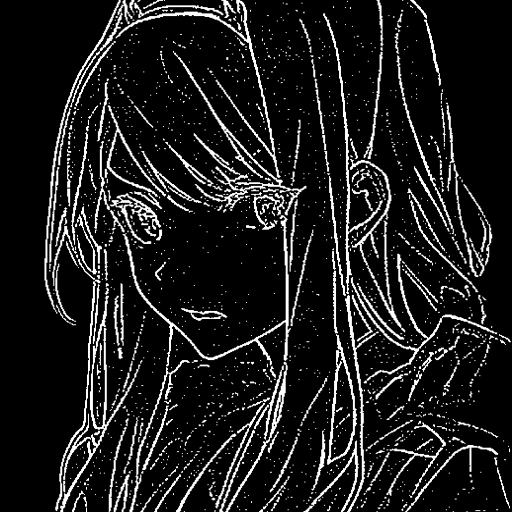}   & \includegraphics[width=\linewidth]{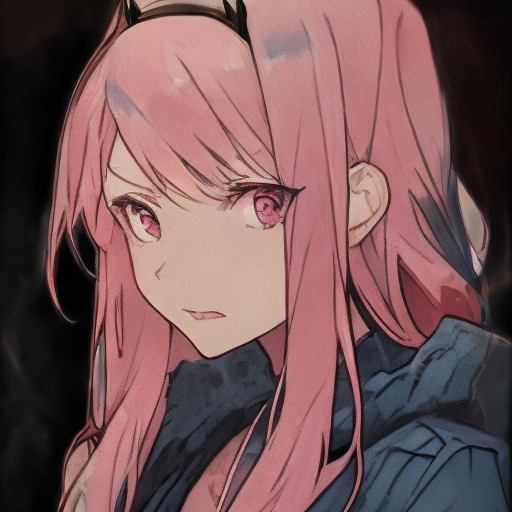}   \\
    \includegraphics[width=\linewidth]{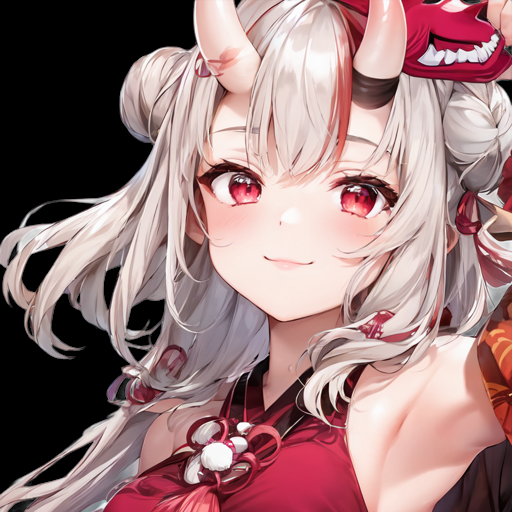}   & \includegraphics[width=\linewidth]{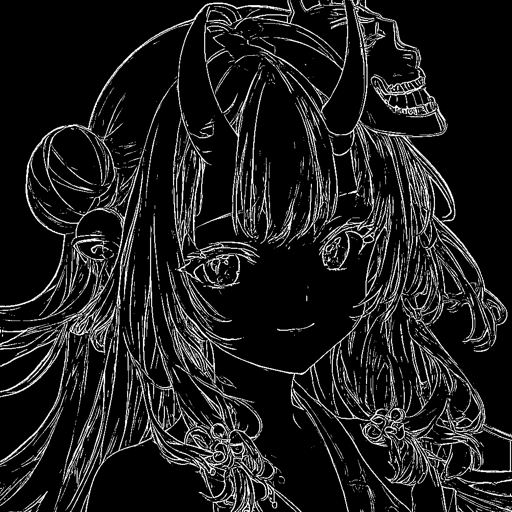}   & \includegraphics[width=\linewidth]{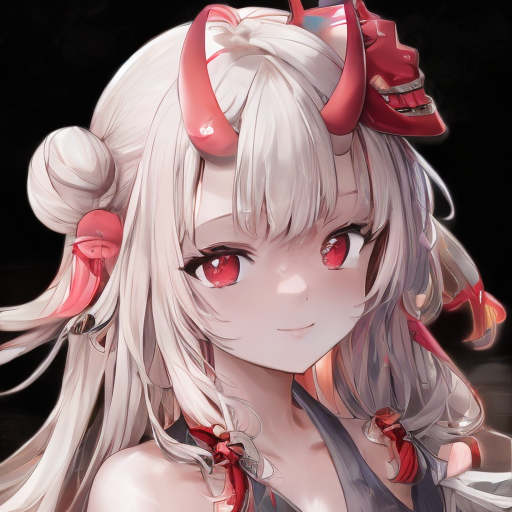}   \\
    \includegraphics[width=\linewidth]{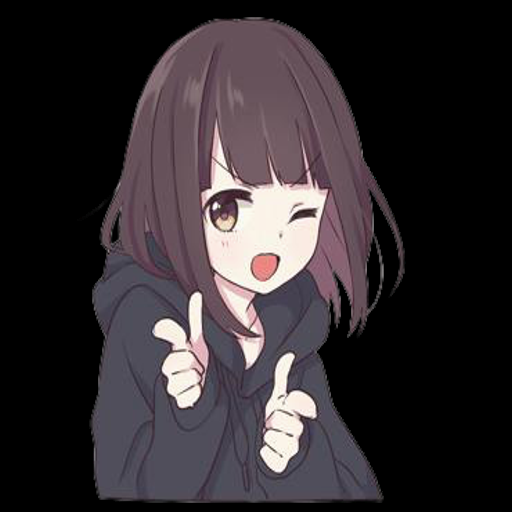}   & \includegraphics[width=\linewidth]{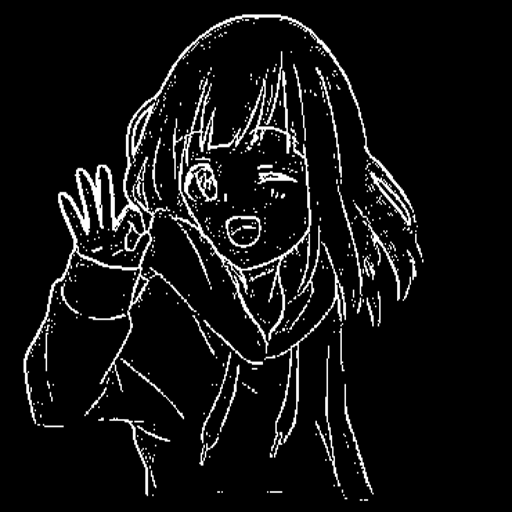}   & \includegraphics[width=\linewidth]{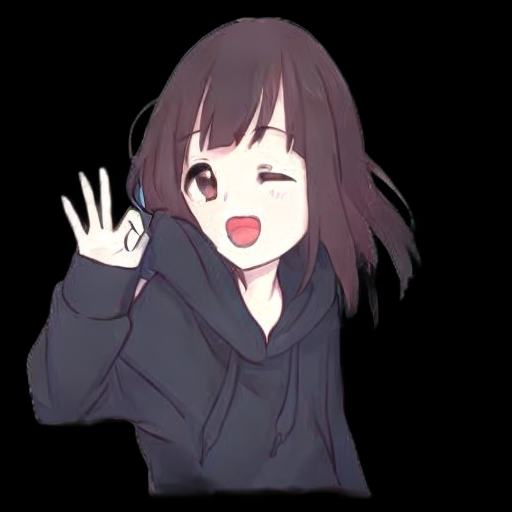}   \\
    \includegraphics[width=\linewidth]{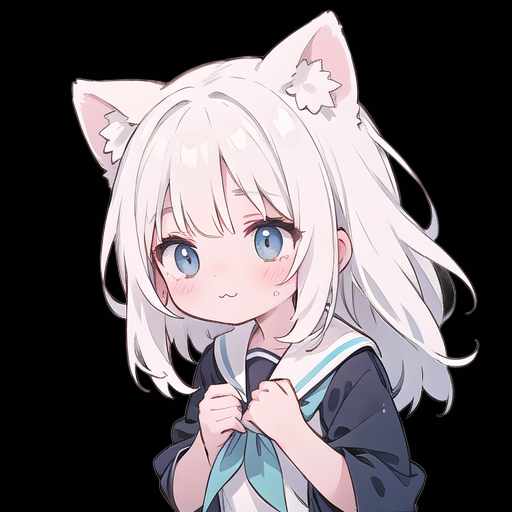}   & \includegraphics[width=\linewidth]{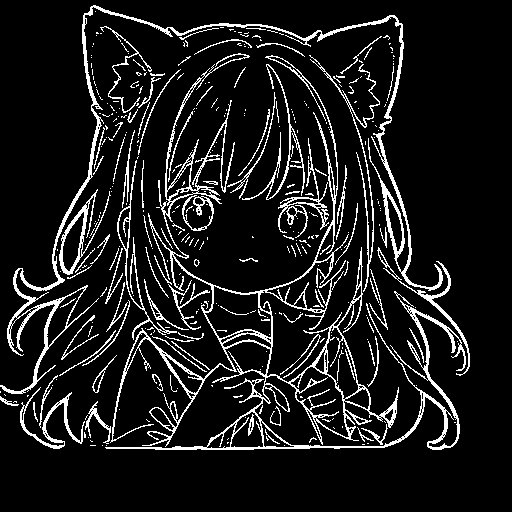}   & \includegraphics[width=\linewidth]{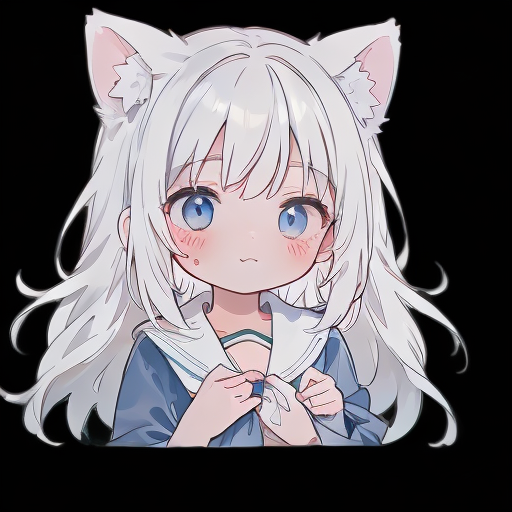}   \\
    \includegraphics[width=\linewidth]{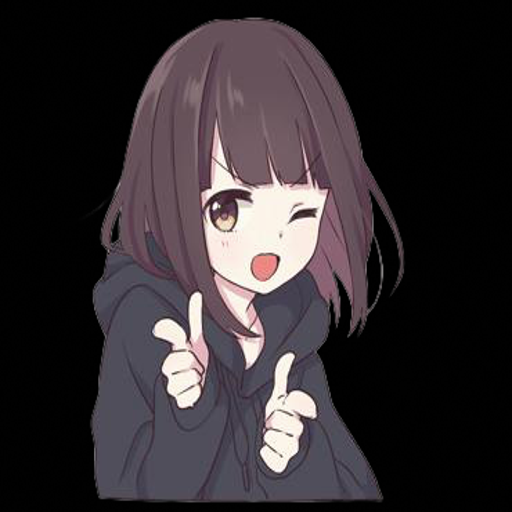} & \includegraphics[width=\linewidth]{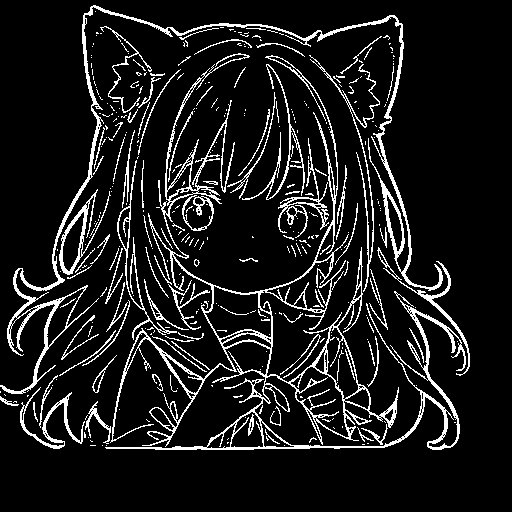} & \includegraphics[width=\linewidth]{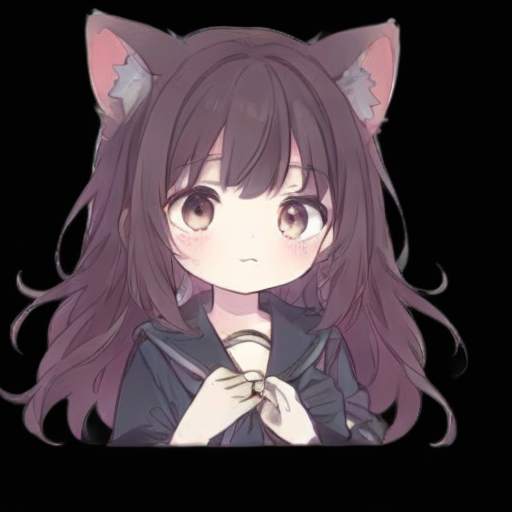} \\
    \bottomrule
  \end{tabular}
  \caption{Stable Diffusion Reference Only: Automatic Coloring Test}
  \label{tab:Stable Diffusion Reference Only: Automatic Coloring Test}
\end{table}

\subsection{Comparison with Existing Methods}

The first comparison method is ControlNet Reference Only. This is a method that
relies on the basic model, so we tested its performance on
stable-diffusion-v1-5-pruned-emaonly and the popular animation-enhanced
AnythingV5 model. ControlNet Reference Only works very poorly with the UniPC
sampler at 20 steps, so we used the configuration parameters it advertises to
work with, the Eluer-a\cite{karras2022elucidating} sampler and sampled at 50
steps. The test environment is stable-diffusion-webui.

The second comparison work is the recently released IP-Adapter, which beats
previous work in related work. We also use the configuration parameters of its
official demo, the DDIM sampler and the number of steps are 50.

Both of the above methods require an additional ControlNet to implement the
line drawing coloring task. We selected ControlNet line art anime and its
corresponding line art processor during training for testing.

The test data is a pair of pictures of the same character or different
characters. In order to avoid possible information differences in blueprints
obtained by different algorithms, the test data includes data oriented to the
coloring of manga characters in real tasks. The comparison method takes the
best result that can be obtained.

The final results are shown in Figure \ref{Comparative Results of Existing
  Methods}. It can be found that even if the animation data of the same style is
just different characters, the performance of the existing methods is quite
different. We think this is caused by whether the UNet it relies on has been
trained with similar images.
\begin{table*}
  \centering
  \begin{tabular}{@{}p{0.11\linewidth}p{0.11\linewidth}p{0.11\linewidth}p{0.11\linewidth}p{0.11\linewidth}p{0.11\linewidth}p{0.11\linewidth}@{}}
    \toprule
    \small{Prompt}                                                       & \small{Blueprint}                                                       & \small{Stable Diffsuion Reference Only}                              & \small{ControlNet \linebreak Reference Only}                                          & \small{ControlNet \linebreak Reference Only + AnythingV5}                                      & \small{IP-Adapter}                                                       & \small{IP-Adapter \linebreak + AnythingV5}                                        \\
    \midrule
    \includegraphics[height=\linewidth,width=\linewidth]{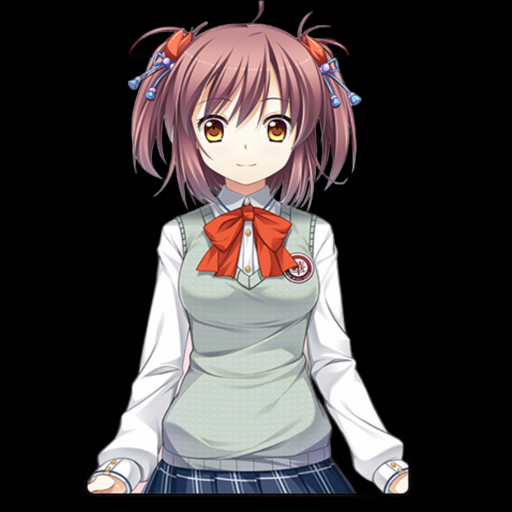}   & \includegraphics[height=\linewidth,width=\linewidth]{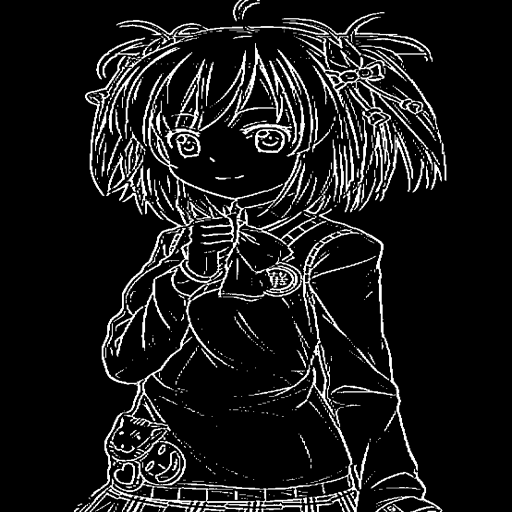}   & \includegraphics[height=\linewidth,width=\linewidth]{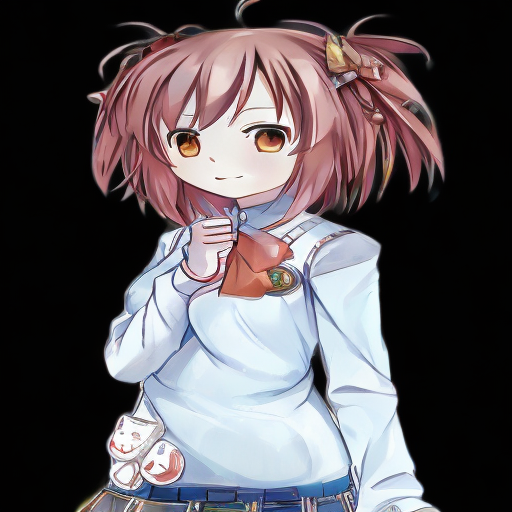}   & \includegraphics[height=\linewidth,width=\linewidth]{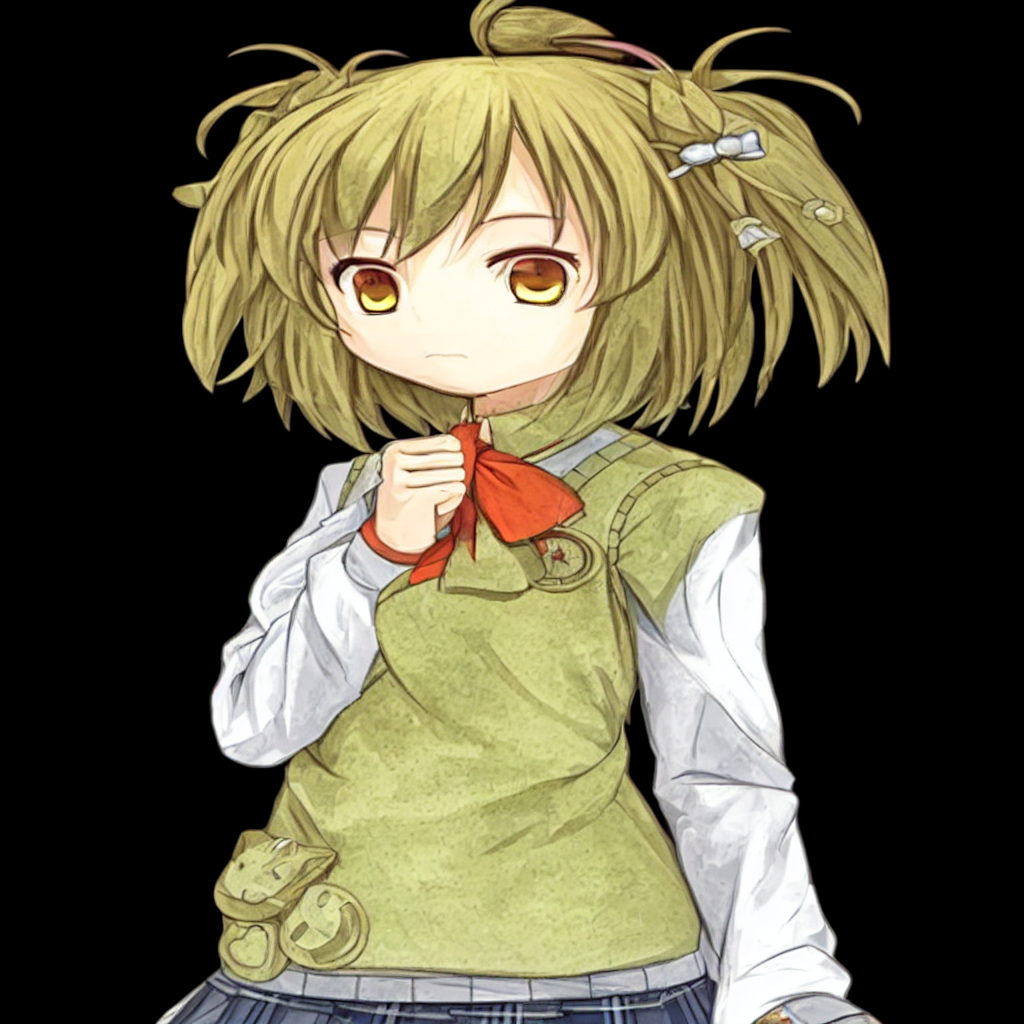 }  & \includegraphics[height=\linewidth,width=\linewidth]{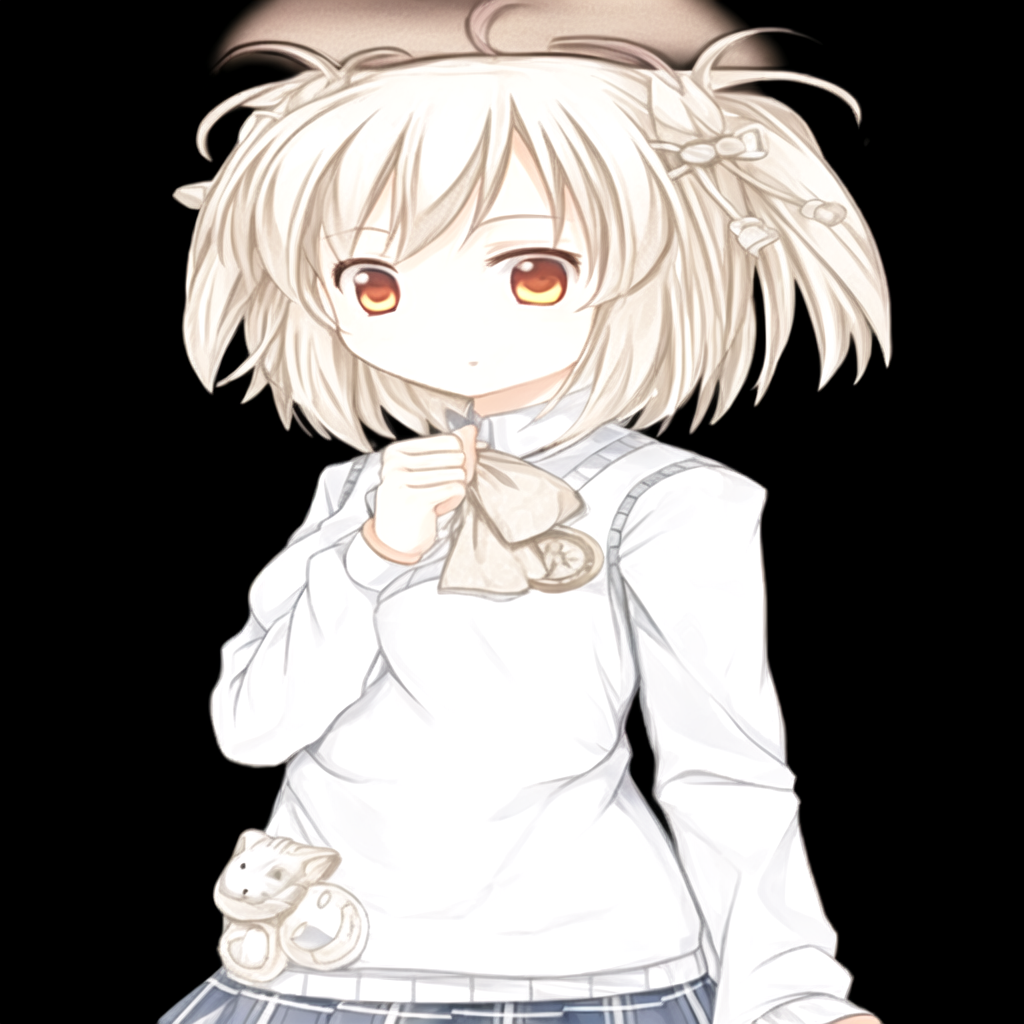}   & \includegraphics[height=\linewidth,width=\linewidth]{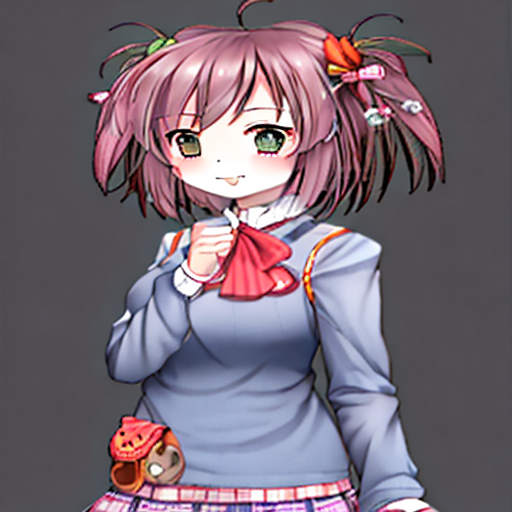}   & \includegraphics[height=\linewidth,width=\linewidth]{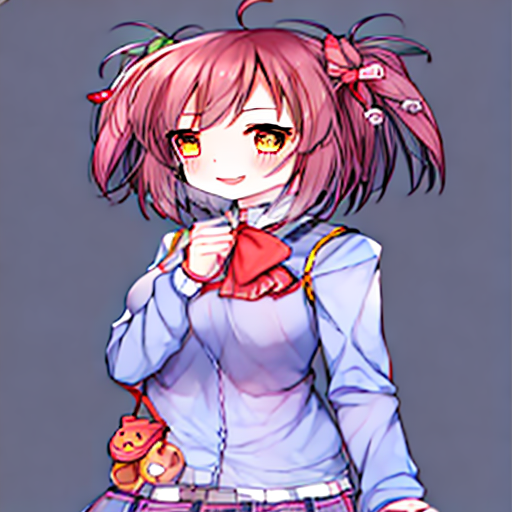}   \\

    \includegraphics[height=\linewidth,width=\linewidth]{2_prompt.png}   & \includegraphics[height=\linewidth,width=\linewidth]{2_blueprint.png}   & \includegraphics[height=\linewidth,width=\linewidth]{2_result.png}   & \includegraphics[height=\linewidth,width=\linewidth]{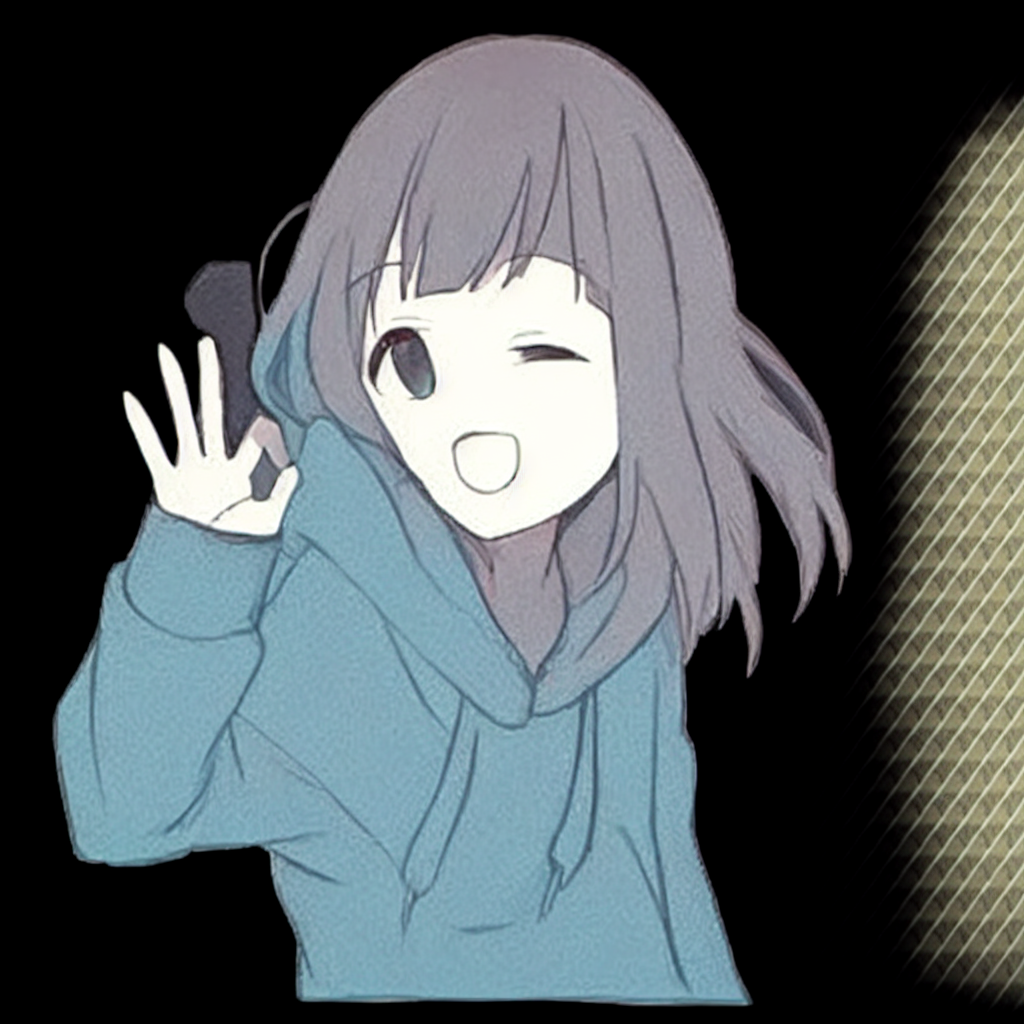}   & \includegraphics[height=\linewidth,width=\linewidth]{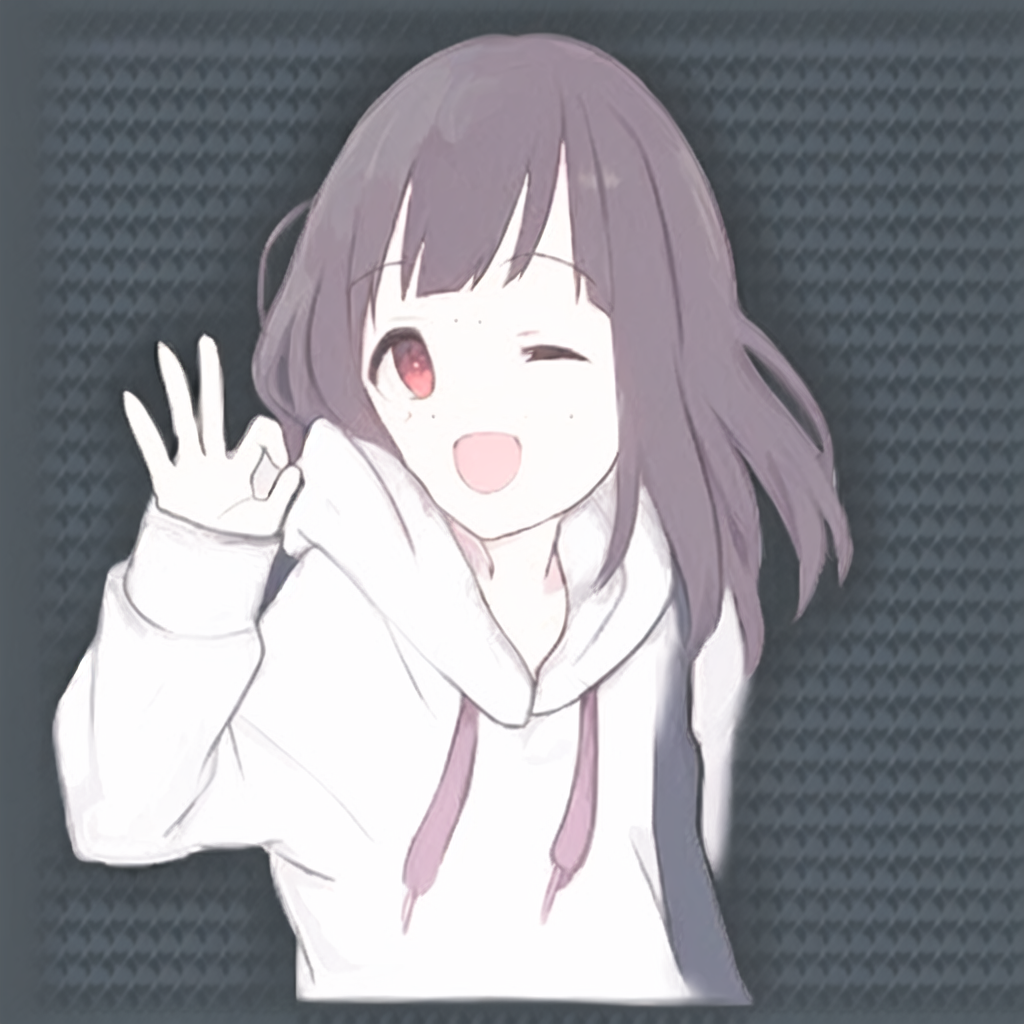}   & \includegraphics[height=\linewidth,width=\linewidth]{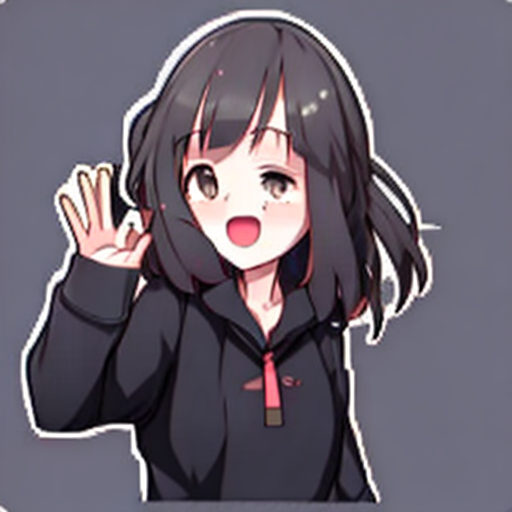}   & \includegraphics[height=\linewidth,width=\linewidth]{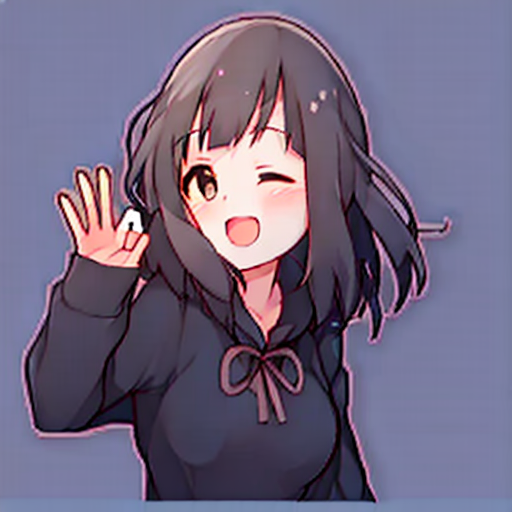}   \\

    \includegraphics[height=\linewidth,width=\linewidth]{3_prompt.png}   & \includegraphics[height=\linewidth,width=\linewidth]{3_blueprint.png}   & \includegraphics[height=\linewidth,width=\linewidth]{3_result.png}   & \includegraphics[height=\linewidth,width=\linewidth]{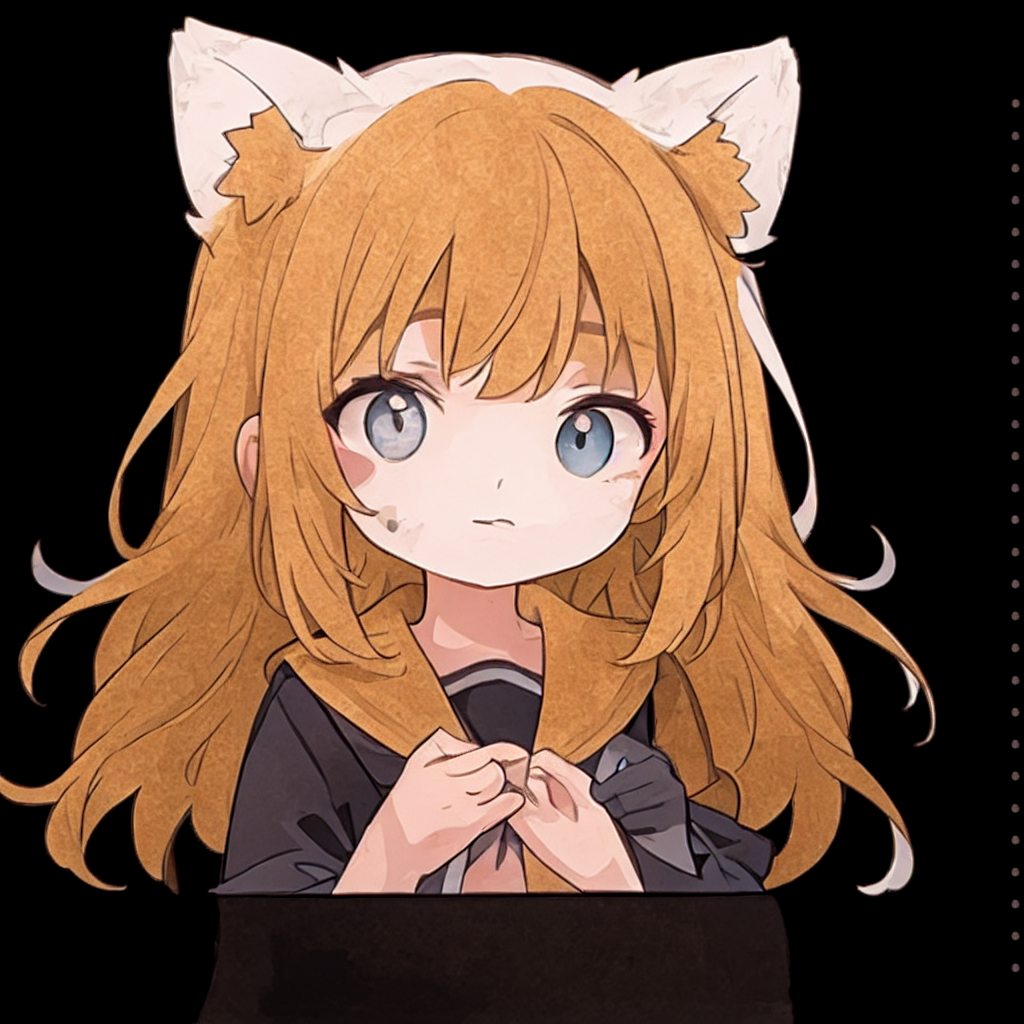}   & \includegraphics[height=\linewidth,width=\linewidth]{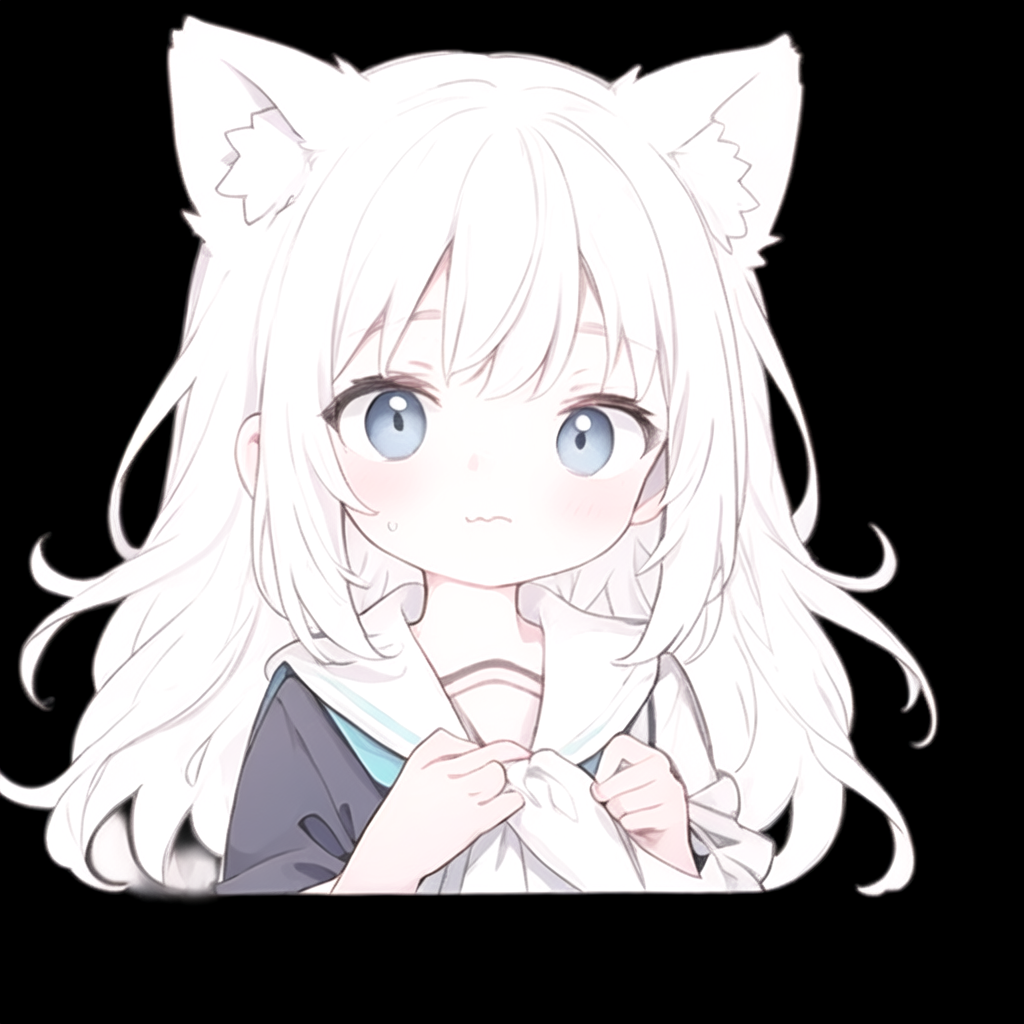}   & \includegraphics[height=\linewidth,width=\linewidth]{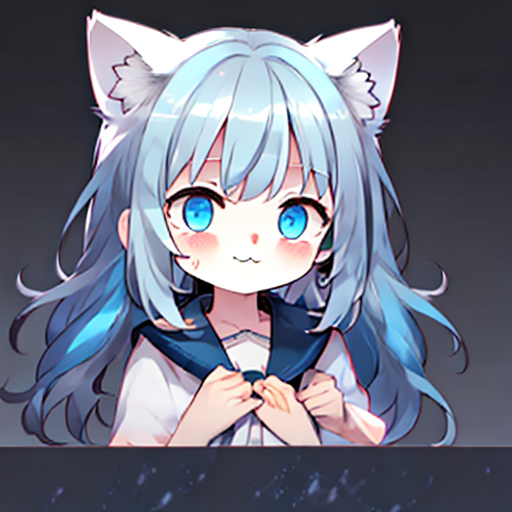}   & \includegraphics[height=\linewidth,width=\linewidth]{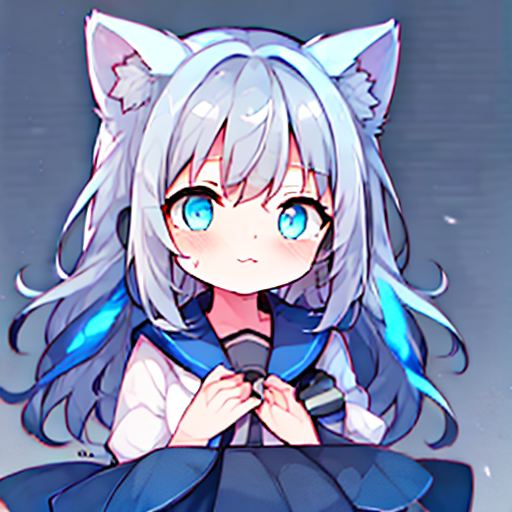}   \\

    \includegraphics[height=\linewidth,width=\linewidth]{4_prompt.png}   & \includegraphics[height=\linewidth,width=\linewidth]{4_blueprint.png}   & \includegraphics[height=\linewidth,width=\linewidth]{4_result.png}   & \includegraphics[height=\linewidth,width=\linewidth]{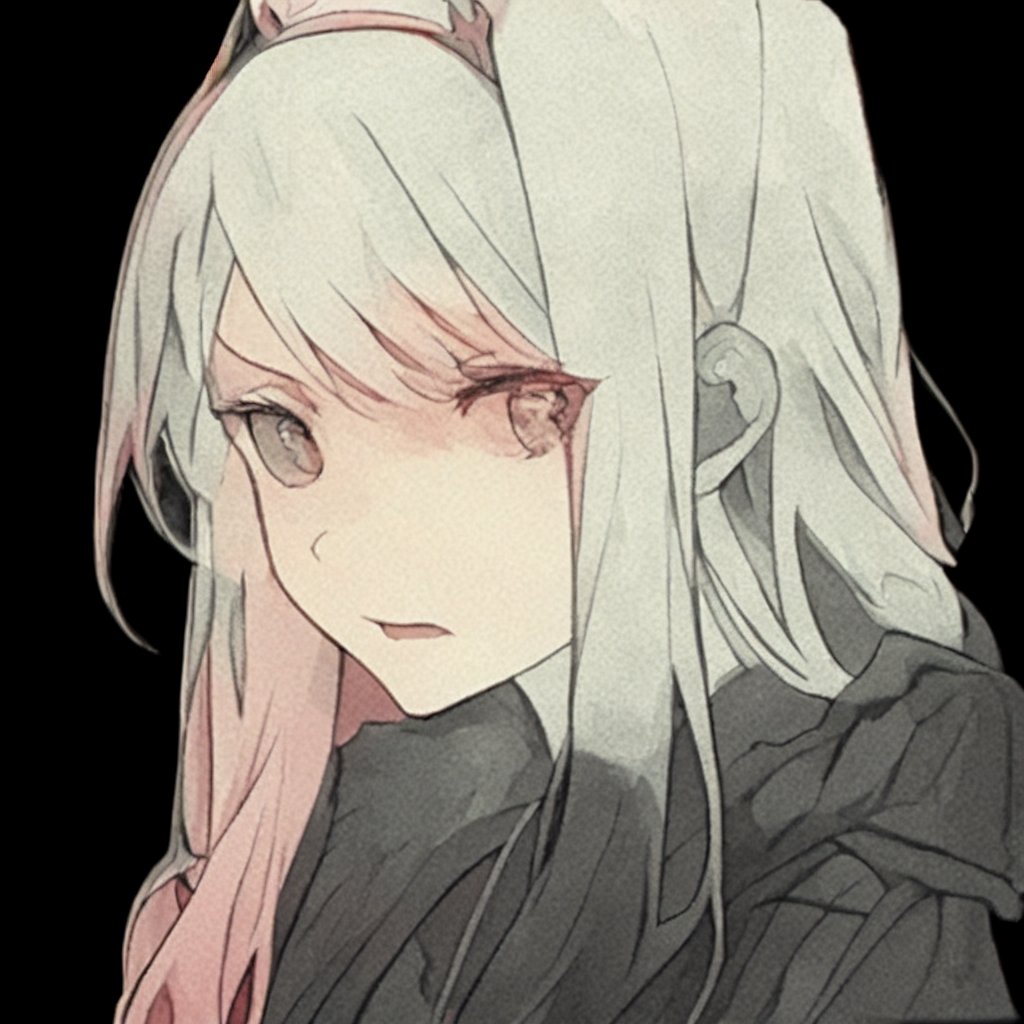}   & \includegraphics[height=\linewidth,width=\linewidth]{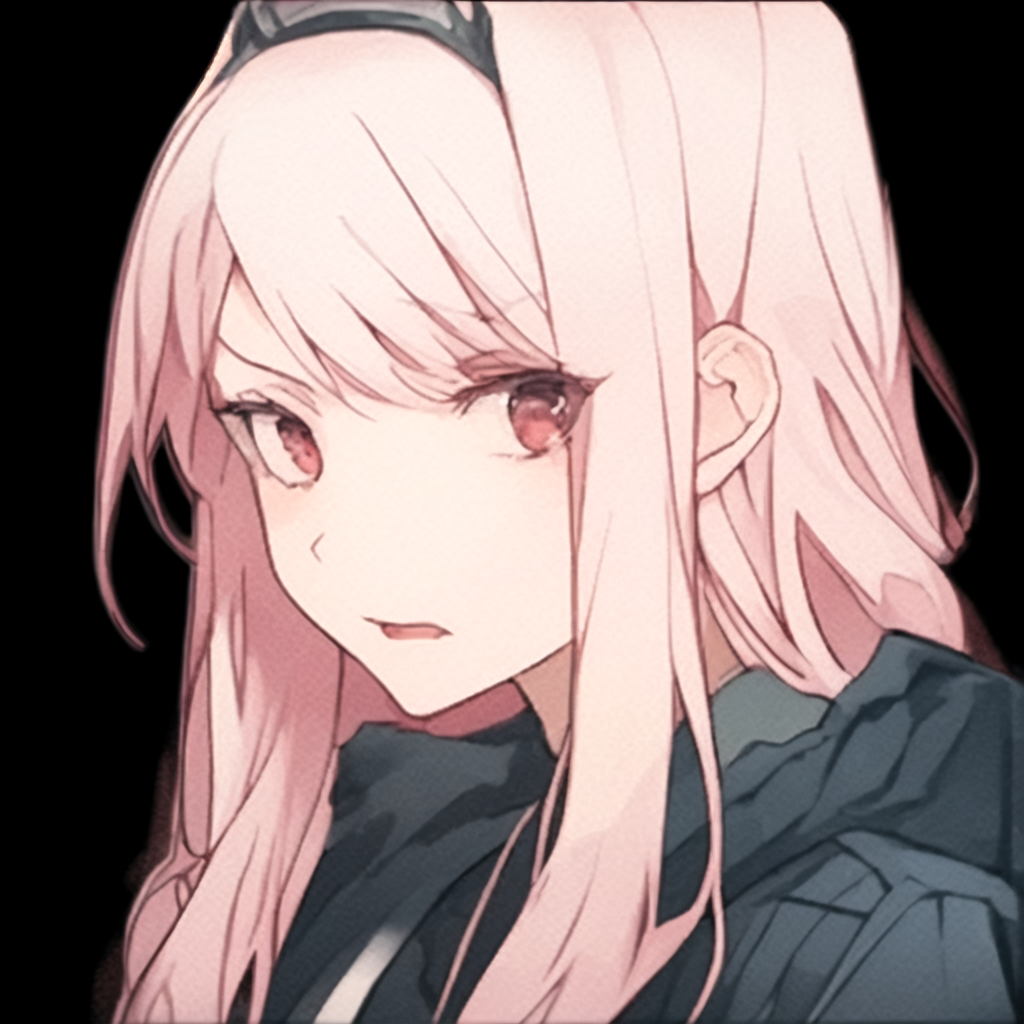}   & \includegraphics[height=\linewidth,width=\linewidth]{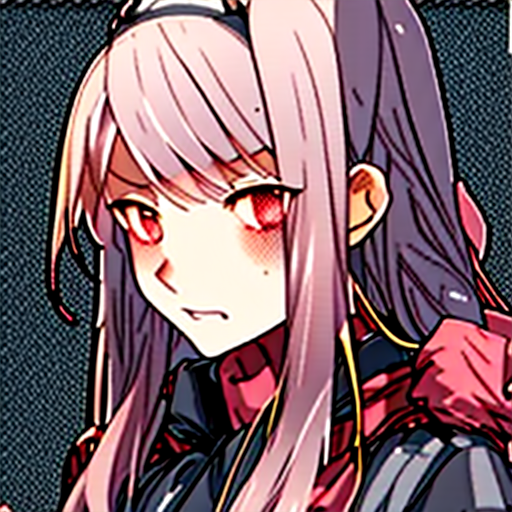}   & \includegraphics[height=\linewidth,width=\linewidth]{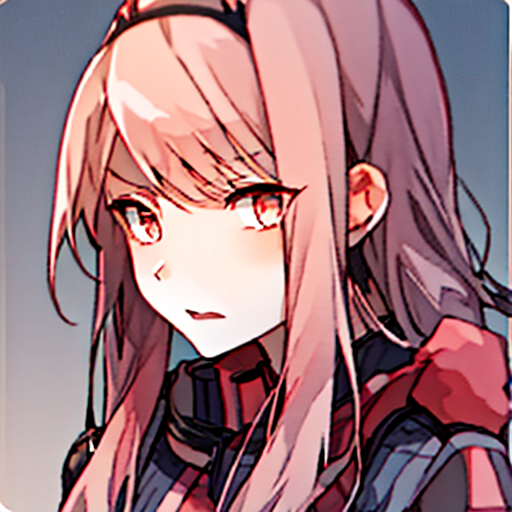}   \\

    \includegraphics[height=\linewidth,width=\linewidth]{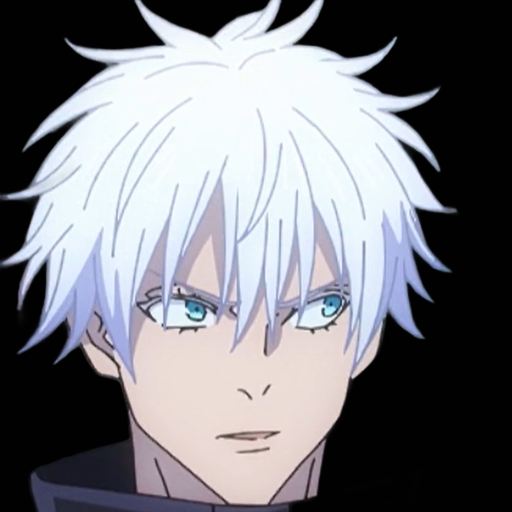}   & \includegraphics[height=\linewidth,width=\linewidth]{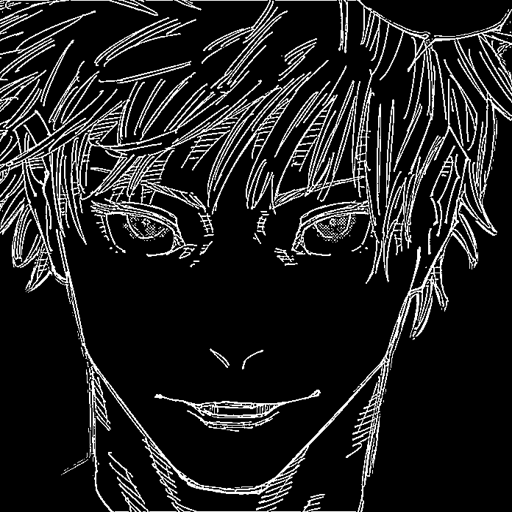}   & \includegraphics[height=\linewidth,width=\linewidth]{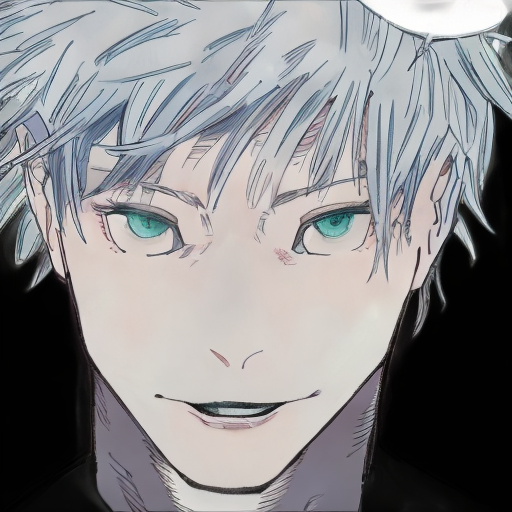}   & \includegraphics[height=\linewidth,width=\linewidth]{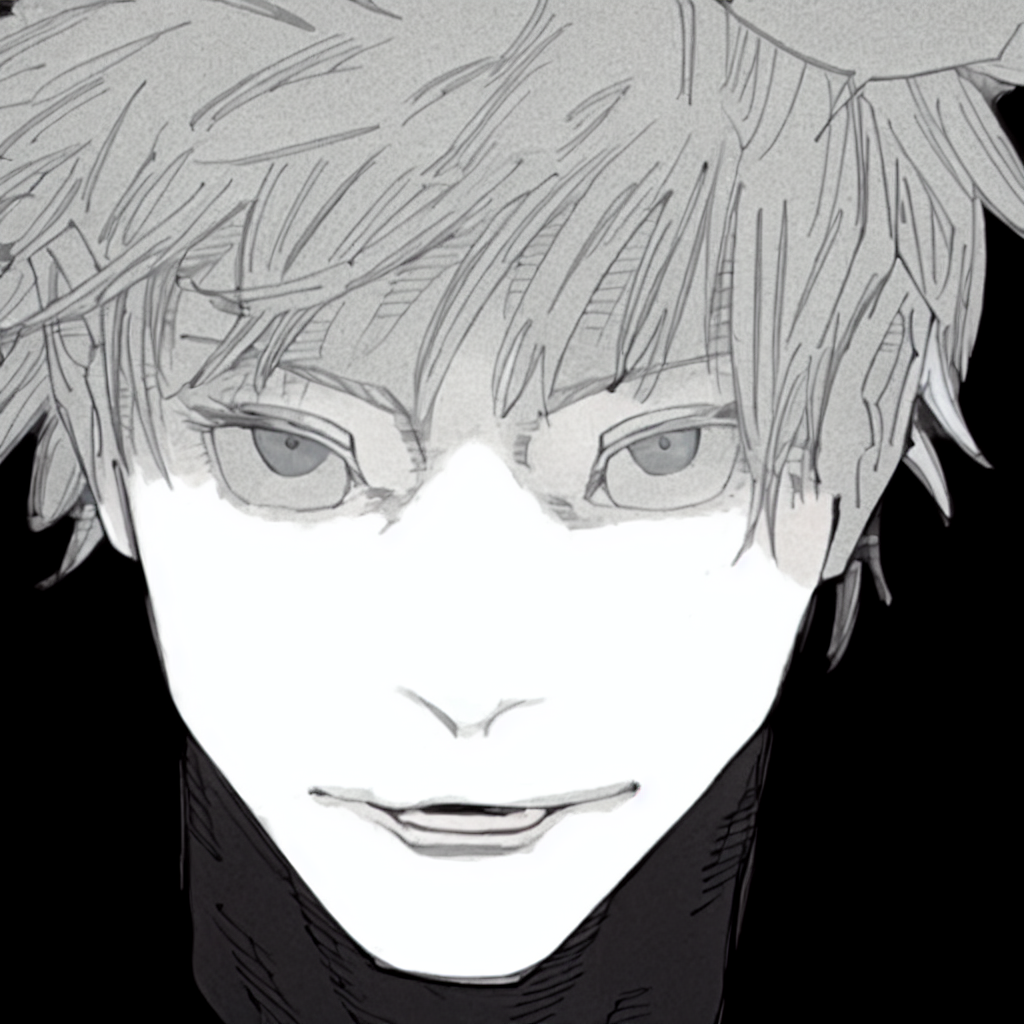}   & \includegraphics[height=\linewidth,width=\linewidth]{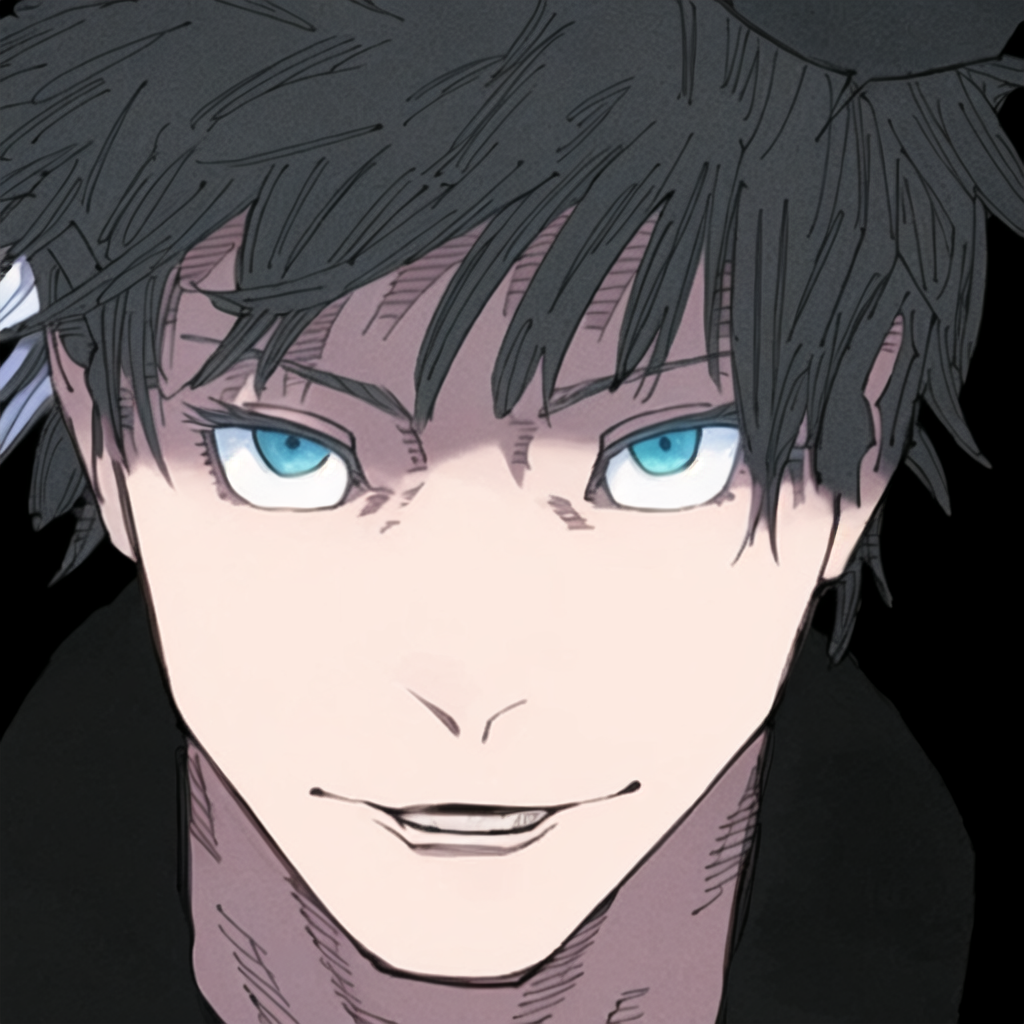}   & \includegraphics[height=\linewidth,width=\linewidth]{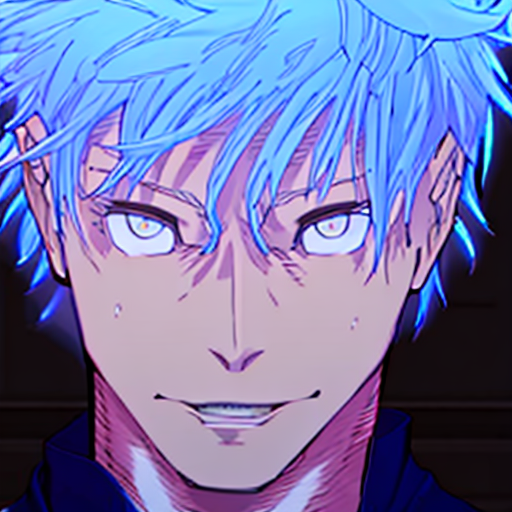}   & \includegraphics[height=\linewidth,width=\linewidth]{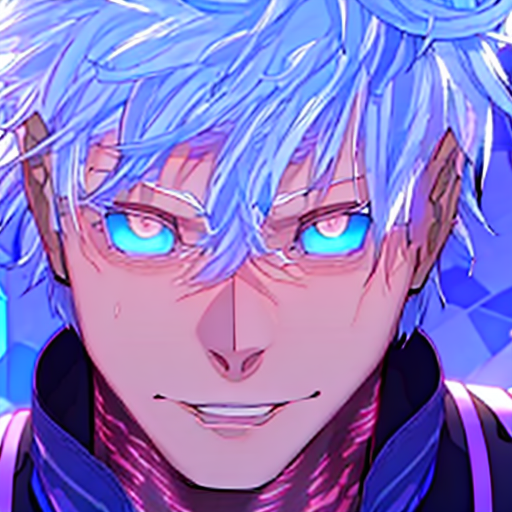}   \\

    \includegraphics[height=\linewidth,width=\linewidth]{6_prompt.png}   & \includegraphics[height=\linewidth,width=\linewidth]{6_blueprint.png}   & \includegraphics[height=\linewidth,width=\linewidth]{6_result.png}   & \includegraphics[height=\linewidth,width=\linewidth]{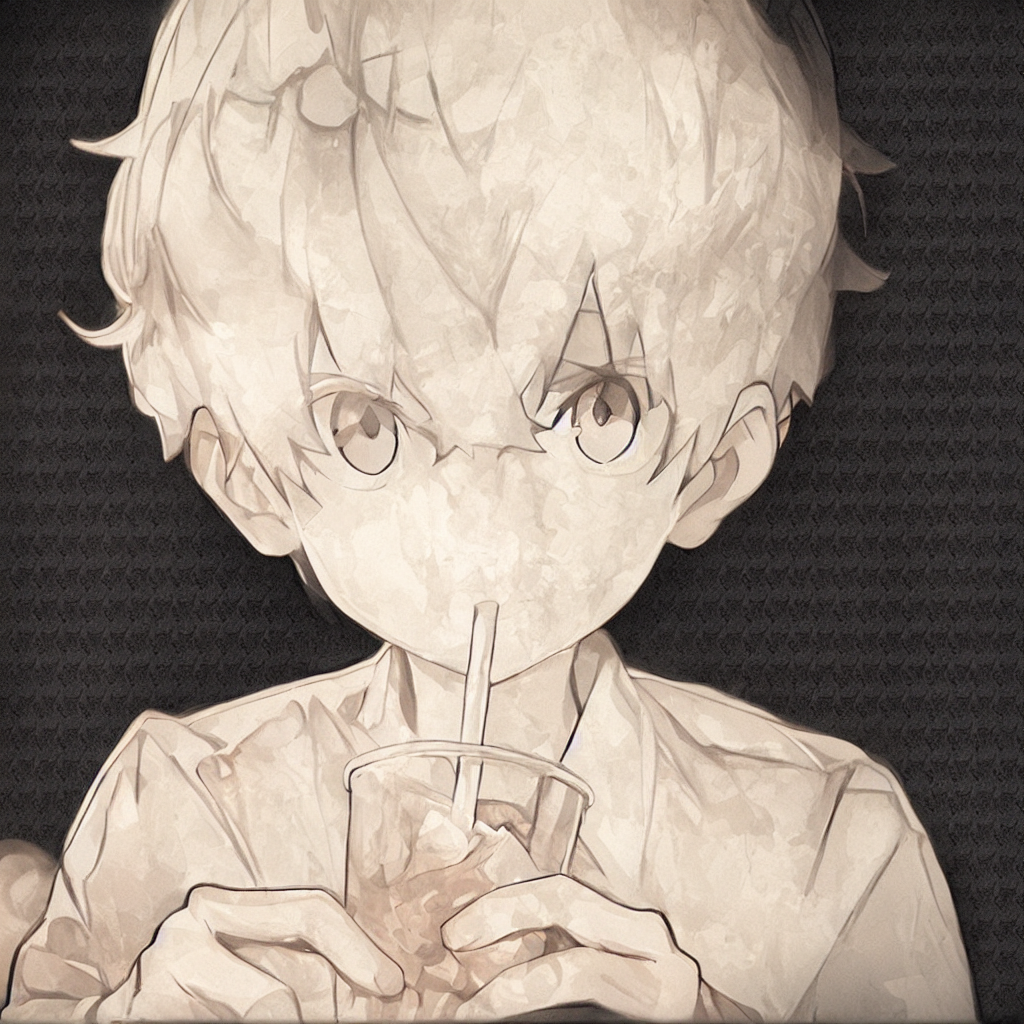}   & \includegraphics[height=\linewidth,width=\linewidth]{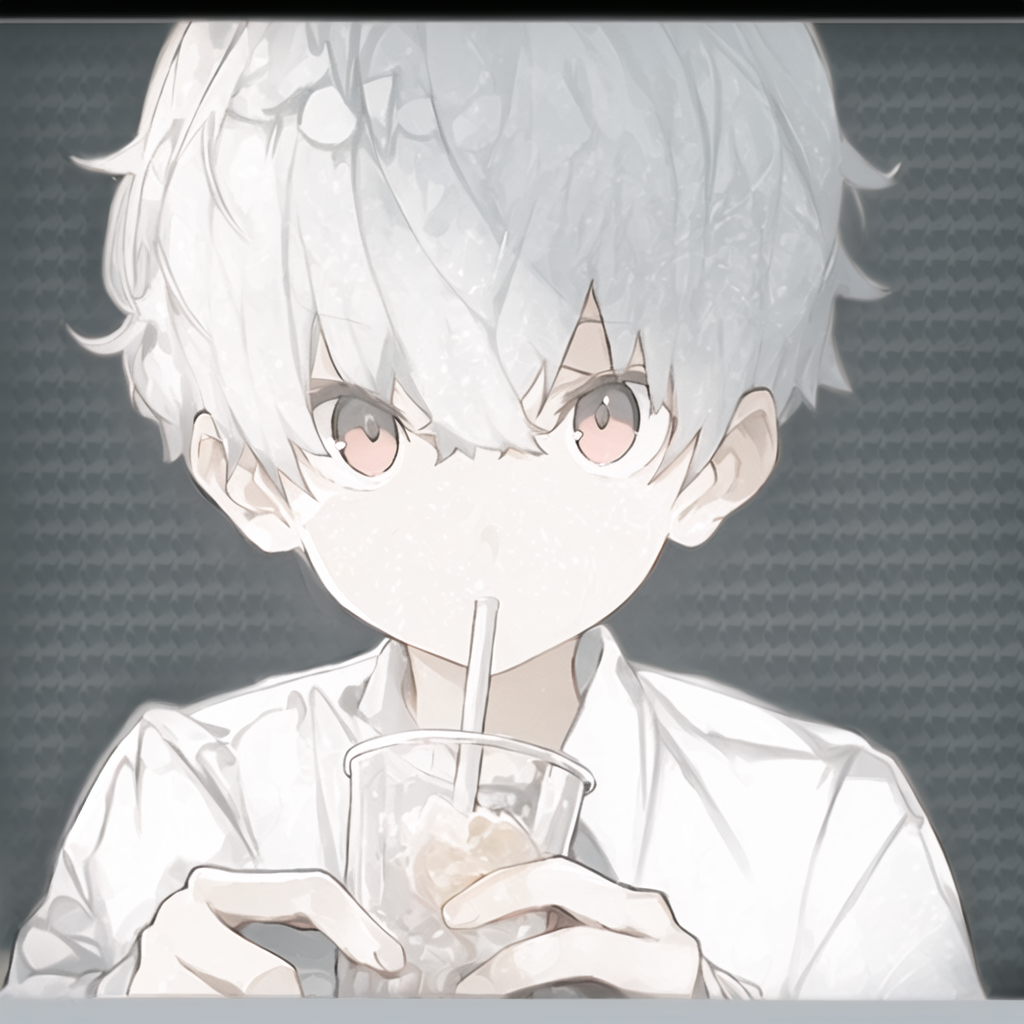}   & \includegraphics[height=\linewidth,width=\linewidth]{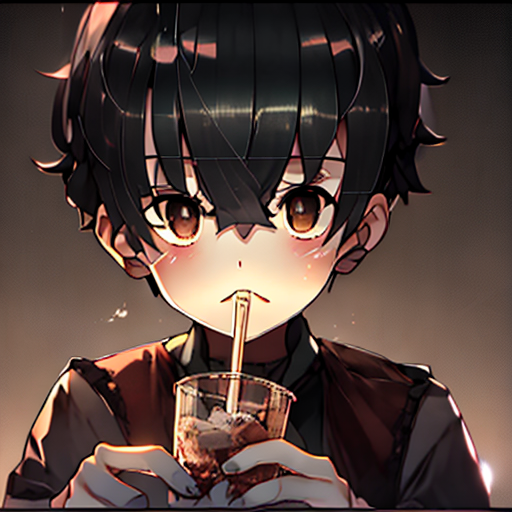}   & \includegraphics[height=\linewidth,width=\linewidth]{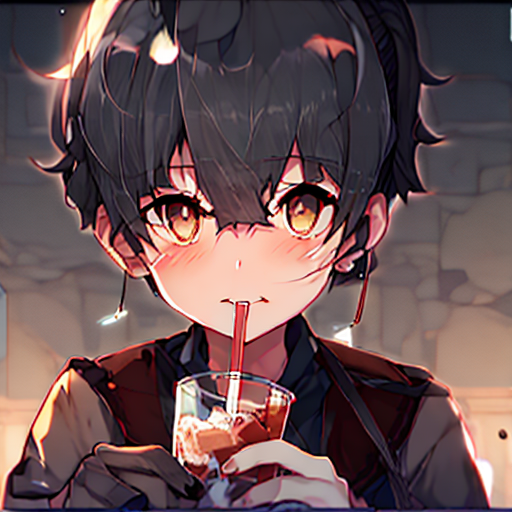}   \\

    \includegraphics[height=\linewidth,width=\linewidth]{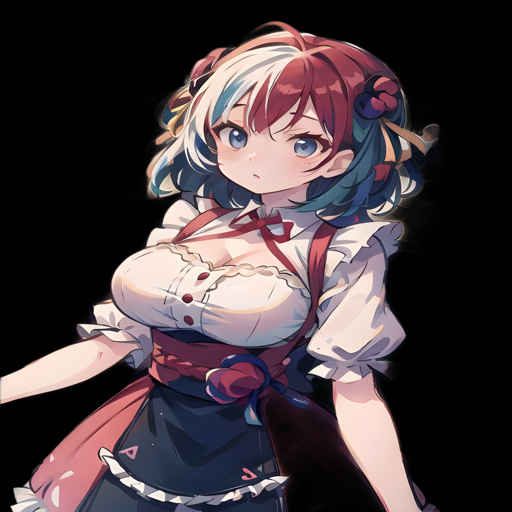}   & \includegraphics[height=\linewidth,width=\linewidth]{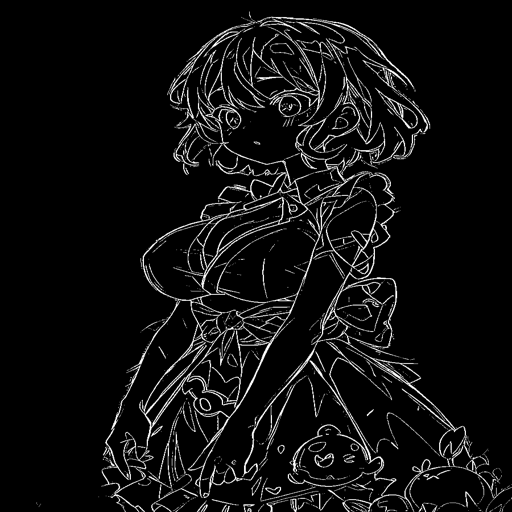}   & \includegraphics[height=\linewidth,width=\linewidth]{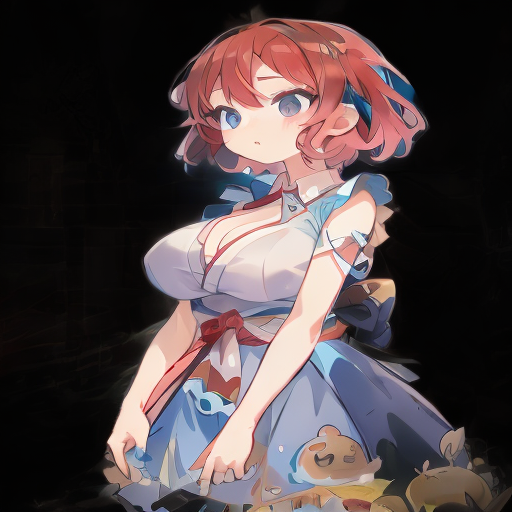}   & \includegraphics[height=\linewidth,width=\linewidth]{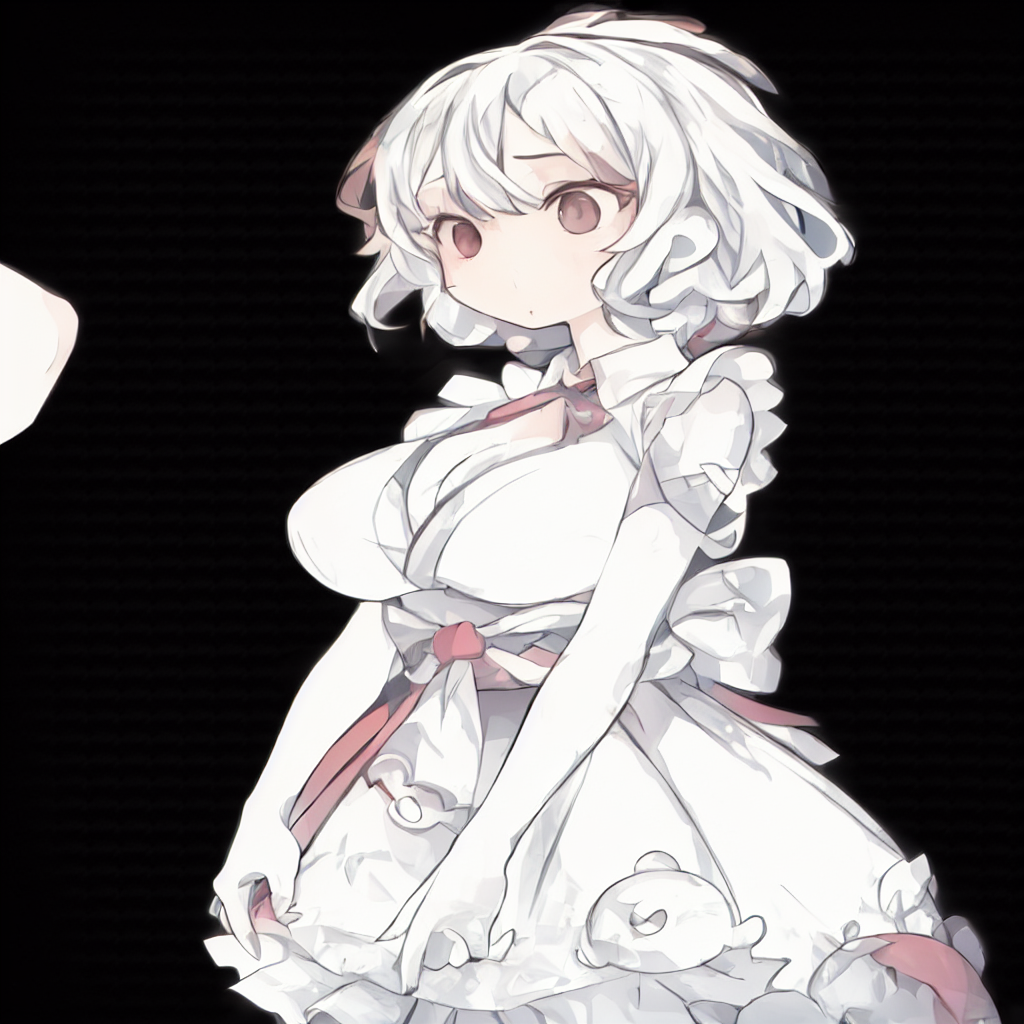}   & \includegraphics[height=\linewidth,width=\linewidth]{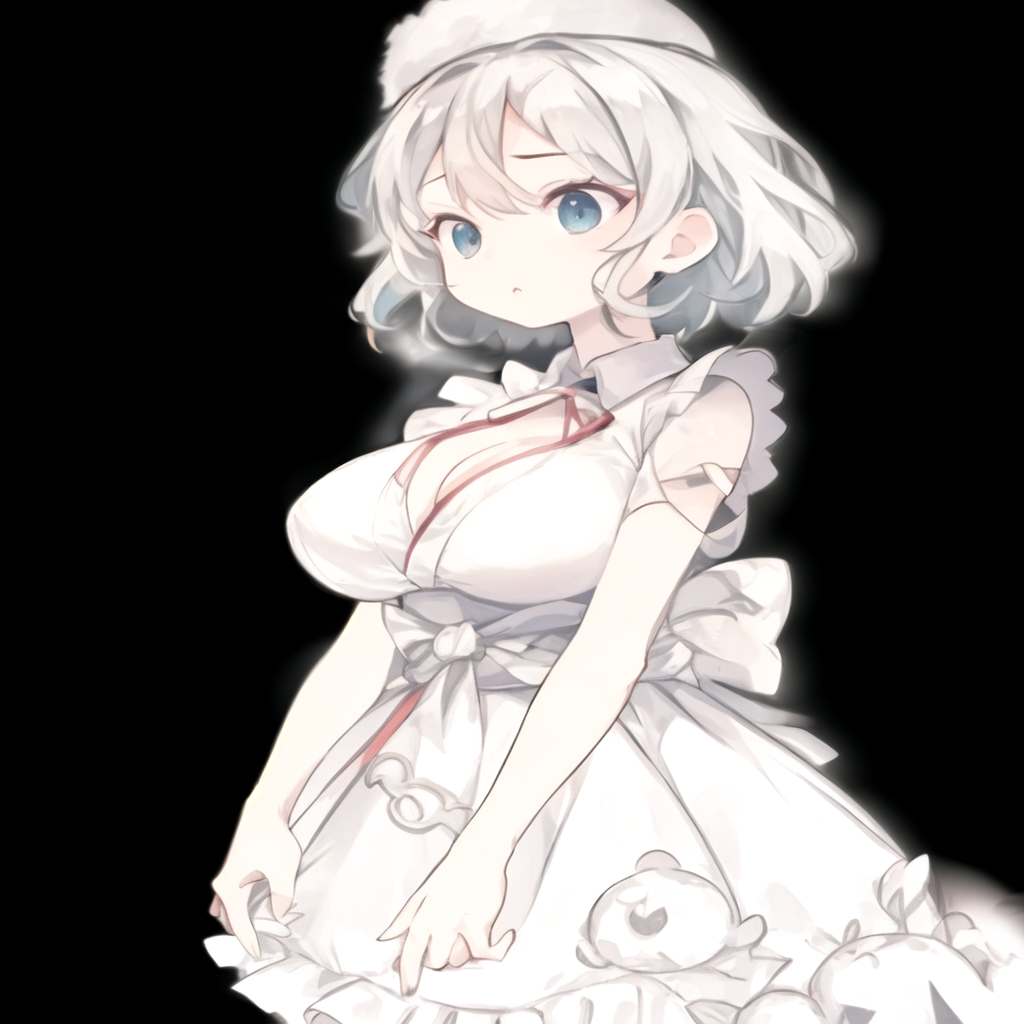}   & \includegraphics[height=\linewidth,width=\linewidth]{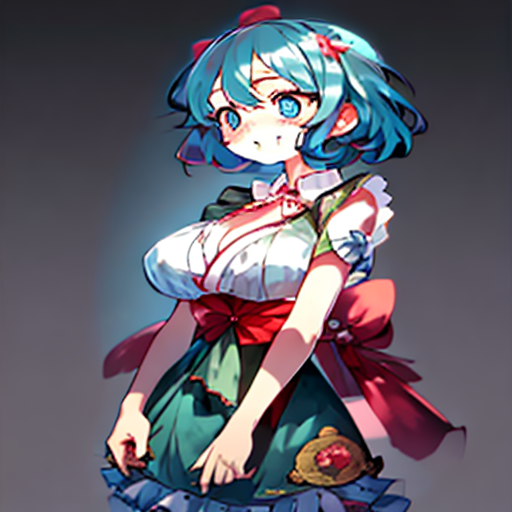}   & \includegraphics[height=\linewidth,width=\linewidth]{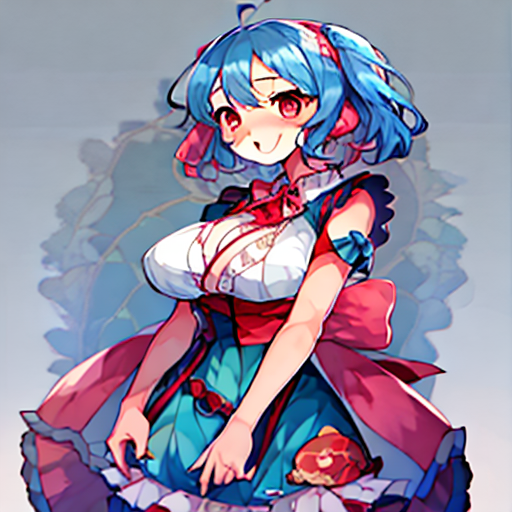}   \\

    \includegraphics[height=\linewidth,width=\linewidth]{2x3_prompt.png} & \includegraphics[height=\linewidth,width=\linewidth]{2x3_blueprint.png} & \includegraphics[height=\linewidth,width=\linewidth]{2x3_result.png} & \includegraphics[height=\linewidth,width=\linewidth]{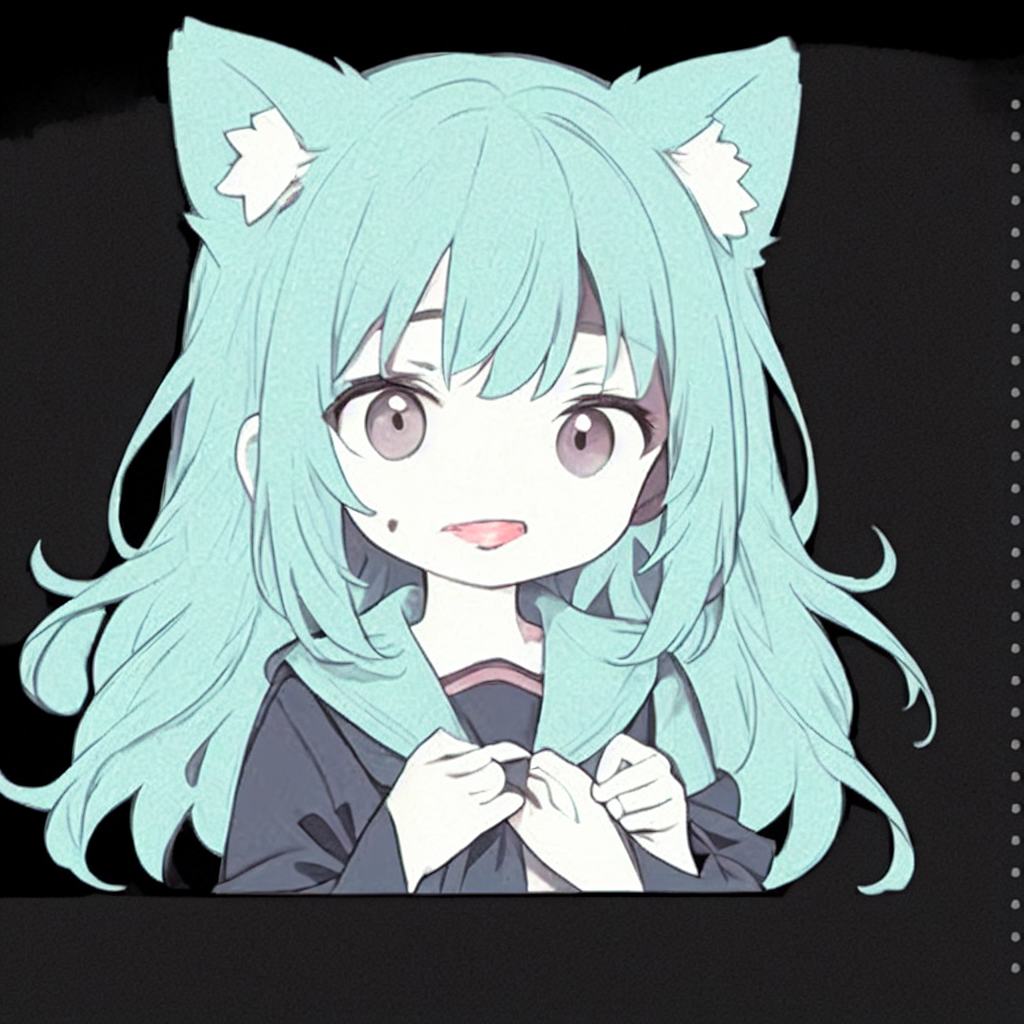} & \includegraphics[height=\linewidth,width=\linewidth]{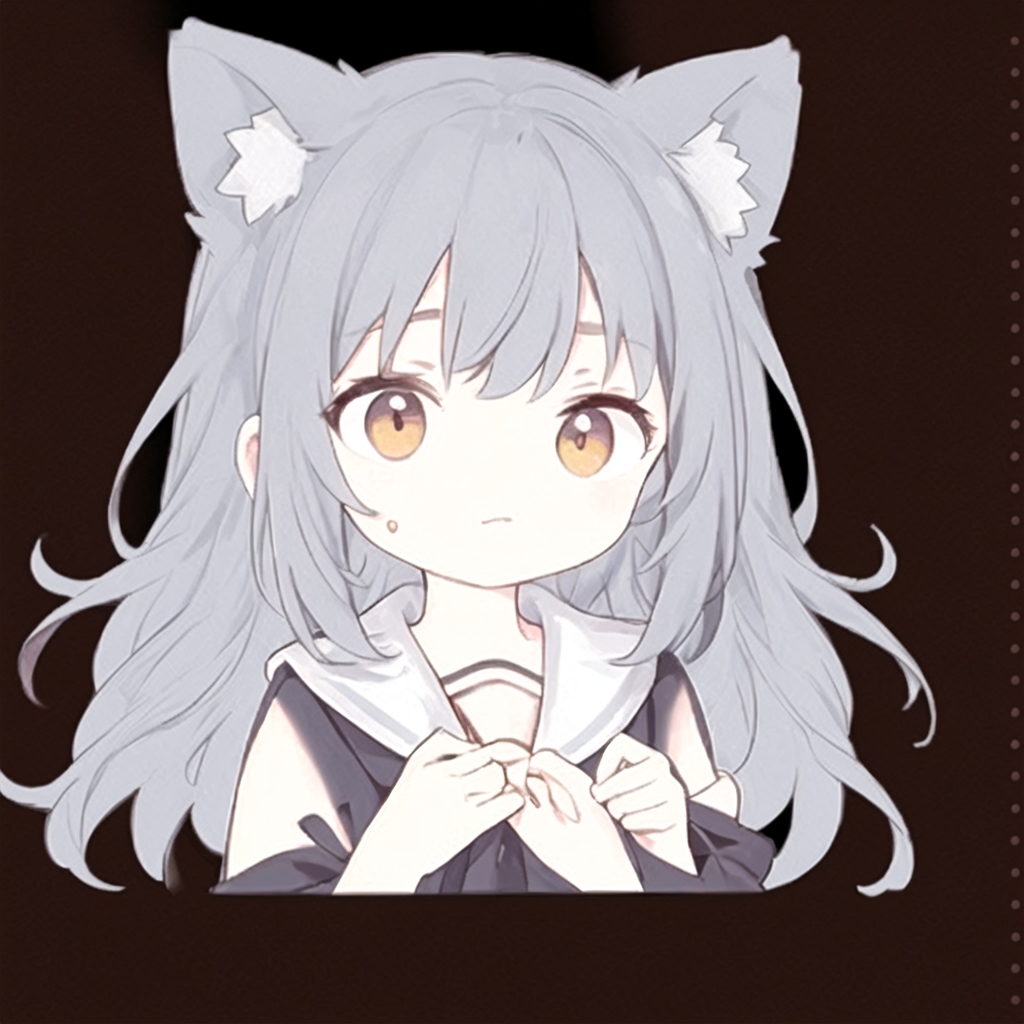} & \includegraphics[height=\linewidth,width=\linewidth]{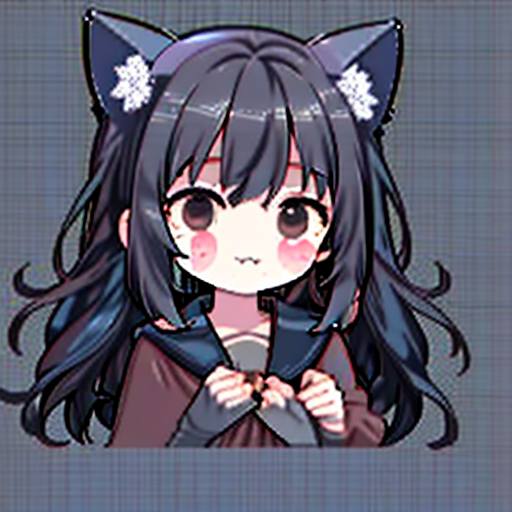} & \includegraphics[height=\linewidth,width=\linewidth]{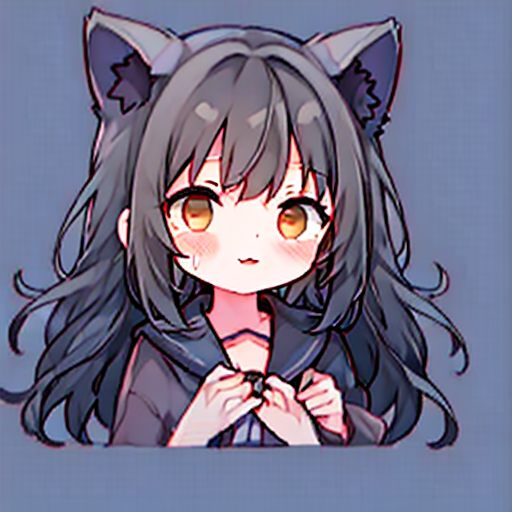} \\
    \bottomrule
  \end{tabular}
  \caption{Comparative Results of Existing Methods}
  \label{Comparative Results of Existing Methods}
\end{table*}

\section{Conclusion}

We have presented image prompt and blueprint jointly guided multi-condition
diffusion model for secondary painting. This is a simple and effective way to
achieve general secondary creation. This joint training achieved by dual
conditions that affect the Q, K, and V of cross attention respectively has
achieved higher generalization and better results in the character line art
coloring task than existing methods. This greatly accelerates the creation
efficiency of animation and comics.

{
    \small
    \bibliographystyle{ieeenat_fullname}
    \bibliography{main}
}


\end{document}